\documentclass[11pt]{article}

\usepackage[final]{acl}

\usepackage{times}
\usepackage{latexsym}

\usepackage[T1]{fontenc}
\usepackage[utf8]{inputenc}

\usepackage{microtype}

\usepackage{inconsolata}

\usepackage{graphicx}

\usepackage{xcolor}

\usepackage{listings}
\usepackage{tcolorbox}
\tcbuselibrary{listings, skins, raster}

\usepackage{amsmath}

\usepackage{float}

\usepackage{subcaption}

\usepackage{tabularx}

\usepackage{booktabs}

\usepackage[utf8]{inputenc}

\title{Improving LLM Code Reasoning via Semantic Equivalence Self-Play with Formal Verification}

\author{Poon Tsz Nok \\
  School of Informatics \\
  University of Edinburgh \\
  \texttt{trevorpoon@gmail.com} \\\And
  Antonio Valerio Miceli Barone \\
  School of Informatics \\
  University of Edinburgh \\
  \texttt{antonio@ed.ac.uk} \\}

\begin{document}
\maketitle

\begin{abstract}
We introduce a self-play framework for semantic equivalence in Haskell, utilizing formal verification to guide adversarial training between a generator and an evaluator. The framework leverages Liquid Haskell proofs for validating equivalence and execution-based counterexamples for inequivalence, organized via a difficulty-aware curriculum. To facilitate this, we release \textbf{OpInstruct-HSx}, a synthetic dataset of $\approx$28k validated Haskell programs. Empirical experiments show that our evaluator transfers effectively to downstream tasks, achieving up to 13.3pp accuracy gain on EquiBench and consistent gains on PySecDB. Ablation studies on the SEQ-SINQ regimes indicate that while inequivalence supervision provides data volume, equivalence proofs are uniquely responsible for the model's reasoning capabilities. The entire training pipeline and dataset are publicly released on GitHub and Hugging Face respectively. 
\end{abstract}

\section{Introduction}

The rise of large language models (LLMs) has reshaped how software can be generated and maintained. Despite models such as Codex and Qwen2.5-coder have demonstrated strong capabilities in producing functional code from natural language prompts~\citep{murphy2024combining, hui2024qwen2}, their outputs often fail to preserve the intended program behavior beyond basic test coverage~\citep{laneve2025assessingcodeunderstandingllms, nguyen2025empirical, wei2025equibench}. This gap raises a fundamental challenge: How can we design training that explicitly teaches models to reason about semantic equivalence between programs? Addressing this problem is critical not only for reliable code generation but also for downstream applications such as program optimization, automated refactoring, and vulnerability detection.

Current approaches largely rely on test suites, which are insufficient for capturing deep semantic properties and edge cases. To bridge this gap, we ground our framework in Haskell for three key reasons. First, its pure functional nature and strong static typing eliminate hidden state and side effects~\citep{thompson2011haskell} (See Appendix~\ref{app:haskell_features} for illustration), making equivalence reasoning mathematically tractable~\citep{launchbury_1993, sestoft_1997}. Second, the Liquid Haskell ecosystem enables the generation of machine-checkable proofs via refinement types, making it possible to certify semantic equivalence for a subset of Haskell programs~\citep{liquidhaskell-intro}. This offers a source of formal supervision that is unavailable in languages such as Python or Java, where formal verification is much harder. Finally, training on underrepresented functional paradigms pushes the model beyond standard object-oriented patterns~\citep{vandam2024investigatingperformancelanguagemodels, giagnorio2025enhancing}, encouraging deeper abstraction capabilities.

We propose a self-play framework for semantic equivalence, in which two specialized agents interact: Alice, a generator that produces variants of reference programs; and Bob, an evaluator trained to decide whether two programs are equivalent. The self-play loop alternates between program generation, verification through proofs or counterexamples, difficulty scoring, and fine-tuning of both agents. By framing the problem as a game between generator and evaluator, the system encourages progressively harder examples and deeper reasoning about semantics.

This work investigates three core questions regarding the utility of functional programming for LLM alignment. First, we examine the dynamics of self-play, asking whether an adversarial loop in a functional language can induce a progressive curriculum that improves semantic reasoning. Second, we evaluate cross-domain and cross-language transferability, assessing whether semantic reasoning skills acquired in Haskell generalize to zero-shot synthesis and vulnerability detection in broader coding benchmarks. Finally, we perform a controlled ablation to determine the relative contributions of supervision signals, distinguishing between the effects of formal equivalence proofs and execution-based counterexamples on evaluator robustness.

\section{Related Work}

Determining semantic equivalence is extremely
challenging in general and current LLMs often fail to recognise semantic equivalence in code~\citep{laneve2025assessingcodeunderstandingllms, nguyen2025empirical, wei2025equibench}. By Rice’s Theorem, determining if two arbitrary programs are semantically equivalent is generally undecidable~\citep{rice1953classes}. Traditional approaches rely on unit testing; however, even extended test suites, such as HumanEval+ and MBPP+ remain insufficient to guarantee correctness~\citep{Gren_2017, Chioteli_2021}. Similarly, symbolic execution offers path-sensitive analysis but suffers from combinatorial state-space explosion as program complexity increases~\citep{badihi2020ardiff}.

To address these incompleteness issues, recent research has pivoted towards formal verification. In the LLM domain, frameworks like DeepSeek-Prover~\citep{xin2024deepseek, xin2024deepseekproverv15harnessingproofassistant, ren2025deepseekproverv2advancingformalmathematical} and Kimina-Prover~\citep{wang2025kimina} have demonstrated that fine-tuning models on self-generated proofs (e.g., in Lean) significantly enhances their verifiable reasoning capabilities. However, generating fully machine-checkable equivalence proofs for imperative languages like Python or C++ is beyond the reach of current tools except for trivial cases~\citep{miceli2025program}.

Our approach bridges this gap by using Haskell and Liquid Haskell. Liquid Haskell embeds refinement types into the language~\citep{Vazou2014LiquidHaskell}, allowing logical properties to be verified automatically via Satisfiability Modulo Theories (SMT) solvers~\citep{Diatchki2015LiquidHaskell, jhala2020programming, liquidhaskell-intro}. This framework enables the construction of machine-checkable lemmas, using reflection and Proof by Logical Evaluation (PLE), to formally certify pointwise equality between candidate functions. This provides a deterministic, high-fidelity feedback signal for the self-play loop.

Self-play has historically driven breakthroughs in game-playing agents like AlphaZero~\citep{silver2017masteringchessshogiselfplay} and OpenAI Five~\citep{openai2019dota2largescale}. In the coding domain, recent frameworks such as Sol-Ver~\citep{lin2025learningsolveverifyselfplay} and AutoIF~\citep{dong2025selfplay} adapt this by using model-generated unit tests and execution feedback to filter synthetic data, achieving significant gains on benchmarks like MBPP and IFEval.

Our work is most directly built upon the adversarial framework proposed by~\citet{miceli2025program}, known as the Semantic Inequivalence Game (SINQ). In their setup, a generator ("Alice") creates program variants and diverging inputs (counterexamples), while an evaluator ("Bob") attempts to detect inequivalence without seeing the generator's justification. This adversarial loop provides a scalable curriculum that improves semantic reasoning. We extend this paradigm by including Semantic Equivalence tasks (SEQ). While~\citet{miceli2025program} focused exclusively on inequivalence via execution feedback, we integrate formal verification using Liquid Haskell. This allows us to train on positive instances of equivalence supported by reasoning traces.

\section{Methodology}

\begin{figure}[htbp]
    \centering
    \includegraphics[width=1.0\linewidth]{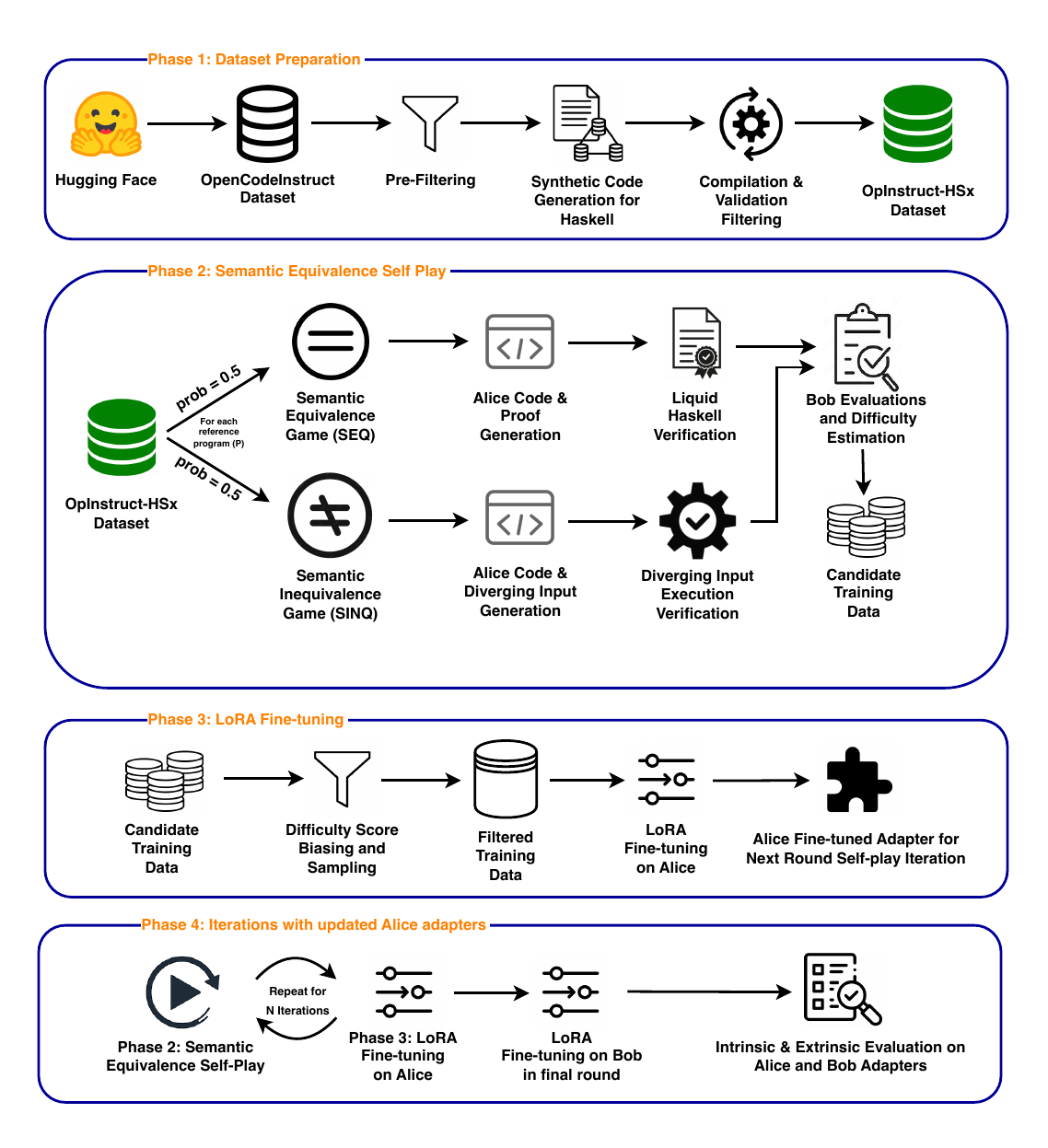}
    \caption{Overview of the semantic self-play framework for improving code reasoning in LLMs via Haskell.}
    \label{fig:self_play_framework}
\end{figure}

We introduce the core methodology of the self-play framework for improving code reasoning in LLMs through two complementary tasks: the SEQ and the SINQ. In the SEQ task, the generator model is asked to produce a functionally identical variant of a given Haskell program, along with a formal proof certifying its equivalence. In contrast, the SINQ task requires generating a function that diverges from the original program on at least one input. Figure~\ref{fig:self_play_framework} shows the complete self-play framework, with difficulty-based supervised fine-tuning that guides the adaptive training loop.

\subsection{Framework Overview}

The self-play framework is formulated as a two-agent adversarial game between Alice (the generator) and Bob (the evaluator). 
\begin{enumerate}
    \item Each round starts with a reference Haskell program $P$, upon which Alice's task is to generate program $Q$, which is either a semantically equivalent variant to $P$ (with a proof) or a semantically inequivalent program (with a diverging input that demonstrates their different behaviour).
    \item The proof or the diverging input is verified afterwards, and if it fails, Alice loses.
    \item Bob then decides whether $(P, Q)$ are semantically equivalent. If Bob’s judgment is accurate, he wins and Alice loses, vice versa.
\end{enumerate}

Alice's objective is to create instances that are difficult for Bob to classify, whereas Bob’s task is to correctly assess them. Through repeated interactions, both agents gradually improve their performance.
\citet{miceli2025program} have proved that this adversarial framework has \textbf{no theoretical upper bound on model's performance improvement}, and in principle both agents can learn endlessly about the complex programming logic while training on a real-world coding dataset.

\subsection{Dataset Generation and Preparation}\label{dataset_prep}

The availability of high-quality Haskell datasets remains extremely limited. Among the few usable resources, the most substantial one is the Blastwind dataset\footnote{\url{https://huggingface.co/datasets/blastwind/github-code-haskell-file}}, which aggregates real-world source files scraped from public GitHub repositories. However, its utility is hindered by substantial noise: the data contains unannotated, inconsistently formatted, and often non-compilable code.

To address the scarcity of high-quality Haskell datasets, we adopt a complementary strategy: \textbf{introducing OpInstruct-HSx, where we generate a synthetic Haskell dataset} by adapting from the \texttt{nvidia-OpenCodeInstruct} dataset~\citep{ahmad2025opencodeinstructlargescaleinstructiontuning}, a large-scale, high-quality instruction corpus originally built for Python code generation.

The programs are first pre-filtered and then transformed into Haskell programs using the \texttt{DeepSeek-R1-Distill-Llama-70B} model. This process results in a synthetic dataset of Haskell programs. To ensure its quality, we applied an automated filtering and validation stage. For each generated Haskell program, we extract the function name and its argument types using syntactic heuristics, and synthesize a type-correct input using a recursive literal generator supporting common base types (e.g., \texttt{Int}, \texttt{Bool}, \texttt{List}, \texttt{Tuple}). Each program is compiled with Glasgow Haskell Compiler (GHC) and executed on the synthesised input. Only those that compile successfully and execute without errors are retained. This filtering process eliminates malformed or non-functional code, ensuring that the resulting dataset consists of minimally functional and executable Haskell programs.

\begin{figure}[htbp]
    \centering
    \includegraphics[width=1.0\linewidth]{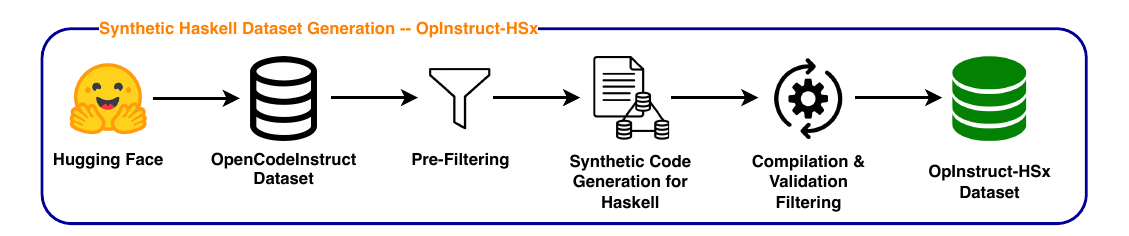}
    \caption{Full Pipeline for the \textbf{OpInstruct-HSx} dataset generation}
    \label{fig:OpInstruct-HSx}
\end{figure}

Figure~\ref{fig:OpInstruct-HSx} shows the entire multi-stage filtering mechanism. We have contributed OpInstruct-HSx, a clean and executable Haskell dataset for both SEQ and SINQ games, which consists of approximately 28,000 validated Haskell functions derived from real-world problems. 
%This dataset is made publicly available at Hugging Face\footnote{\url{https://huggingface.co/datasets/Trevor0501/OpInstruct-HSx}}, serving as a high-quality synthetic Haskell resource for training LLMs in semantic reasoning tasks.
This dataset is made publicly available\footnote{\url{https://huggingface.co/datasets/Trevor0501/OpInstruct-HSx}}, serving as a high-quality synthetic Haskell resource for training LLMs in semantic reasoning tasks.
The code to create the data and replicate the experiment is released on a public repository\footnote{\url{https://github.com/TrevorPoon/llm-self-play-liquidhaskell}}. 
%\footnote{\url{https://anonymous.4open.science/r/llm-self-play-liquidhaskell-10C4}}

\subsection{The Self-Play Loop: Alice and Bob}

\subsubsection*{Step 1: Program Selection and Branching}\label{sec:branching}

Let $\mathcal{D}$ be the dataset of reference Haskell programs. We randomly choose a reference program $P \in \mathcal{D}$ and then select the SEQ game with 50\% probability, otherwise we choose the SINQ game. %This setup ensures equal chance for both SEQ and SINQ challenges in training.

\subsubsection*{Step 2a: SEQ Game (Alice’s Turn)}

In the SEQ game, Alice receives $P$ and must synthesize $Q$ such that for all inputs $x$:
$$
\forall x \in \mathcal{X}.\; P(x) = Q(x)
$$
To challenge Bob, Alice is encouraged to construct a hard instance $Q$, aiming for a maximum target difficulty level (d=10, defined in Section \ref{SEC:METHOD:IMPL:DIFFICULTY}). In addition, Alice is required to produce a formal proof (in Liquid Haskell) of this semantic equivalence. See Appendix~\ref{app:LH_instances} for an SEQ instance.

\subsubsection*{Step 2b: SINQ Game (Alice’s Turn)}

Alternatively in the SINQ game, Alice is instructed to produce a function $Q$ that diverges from $P$ on at least one input:
$$
\exists x^* \in \mathcal{X}: P(x^*) \neq Q(x^*)
$$
Alice is also incentivized to construct a difficult function $Q$ that Bob is likely to misclassify, again targeting a maximum difficulty level (d=10). Alice must also output a diverging input $x_a$ showing this inequivalence such that $P(x_a) \neq Q(x_a)$. See Appendix~\ref{app:LH_instances} for an SINQ instance.

\subsubsection*{Step 3: Verification through Liquid Haskell or Execution}

\begin{itemize}
    \item \textbf{SEQ Game}: Alice’s proof is verified by Liquid Haskell, which acts as an external oracle. If the proof is accepted, the proof is retained as a fine-tuning example.
    \item \textbf{SINQ Game}: The candidate $x_a$ is tested by an execution: If $P(x_a) \neq Q(x_a)$, the instance is accepted. Otherwise, the sample is discarded.
\end{itemize}

All candidates undergo compilation, execution, and formal verification checks. This ensures the training data for both Alice and Bob remains high-quality and executable.

\subsubsection*{Step 4: Bob’s Turn -- Difficulty Estimation}

After Alice produces her candidate program $Q$, Bob is presented with both $P$ and $Q$ only and must decide whether the two programs are semantically equivalent. To estimate the challenge posed by each example, Bob is sampled $N$ times and the proportion of correct responses from Bob is then used to compute a difficulty score. Further details on the difficulty-based curriculum and dataset sampling can be found in Section~\ref{SFT_w_diff}.

\subsection{Implementation with Supervised Fine‐Tuning with Difficulty Score}
\label{SFT_w_diff}

Given the practical challenges of reinforcement learning (Appendix~\ref{app:RL}), we instead adopt rejection sampling supervised fine-tuning (SFT). We construct fine-tuning datasets by having Alice generate challenging programs and Bob learn from his own correct identifications. This semi-adversarial pipeline ensures Alice continually refines her ability to craft borderline-difficult SEQ or SINQ program pairs, while Bob steadily improves at its reasoning ability in semantic equivalence. Detailed prompt formats for both agents are provided in Appendix~\ref{both_prompt}.

\subsubsection{Alice's Training Data Selection}
\label{SEC:METHOD:IMPL:DIFFICULTY}

After Alice generates the candidate pairs $(P, Q)$, Alice's outputs are not immediately used for fine-tuning. Instead, for each pair, Bob is asked to evaluate their semantic relation multiple times (typically $N=10$). The number of correct Bob responses $N_\text{success}$ determines the \textbf{difficulty score} $\hat d$ as defined in Equation~\ref{Bob_difficulty_level}.
\begin{equation}\label{Bob_difficulty_level}
    \hat d = d(P,Q) = 10 \times \left(1 - \frac{N_\text{success}}{N}\right)    
\end{equation}
However, most of Alice’s early generations are trivial for Bob. Including all examples would flood the dataset with low-difficulty cases that Bob already solves easily. Following the design in Miceli-Barone et al.~\citep{miceli2025program}, we only retain examples that are sufficiently challenging, as determined by the difficulty score $\hat d$. Hence, shown in Schema~\ref{training_data_split}, we split all $(P,Q)$ pairs into $\mathcal{D}_{Hard}$ and $\mathcal{D}_{Easy}$, where $\tau$ is a chosen difficulty threshold (e.g., $\tau = 5$). 
\begin{equation}\label{training_data_split}
\begin{split}
    \mathcal{D}_{Hard} &= \left\{ (P, Q) \;\middle|\; d(P, Q) > \tau \right\}, \\ 
    \mathcal{D}_{Easy} &= \left\{ (P, Q) \;\middle|\; d(P, Q) \leq \tau \right\}
\end{split}
\end{equation}

Finally, Alice’s training dataset is comprised of three SFT examples for each validated pairs $(P,Q)$. The first example pairs the prompts and Alice's generation, and directly trains Alice to generate a challenging program $Q$, which helps improve Alice’s ability to craft precise SEQ or SINQ code (Schema~\ref{Alice_first_example}). $\mathrm{SP}$ and $\mathrm{UP}$ denote the system and user prompts conditioned on the reference program $P$ and a target difficulty $d=10$. $O$ represents the model's raw output, which includes the chain-of-thought followed by the generated program $Q$ (and the diverging input $x$ in the case of SINQ). 
\begin{equation}\label{Alice_first_example}
\begin{split}
    \text{SEQ} &= \bigl(\mathrm{SP}_A^{eq},\; \mathrm{UP}_A^{eq}(P,10),\; O_A^{eq} \bigr) \\
    \text{SINQ} &= \bigl(\mathrm{SP}_A^{inq},\; \mathrm{UP}_A^{inq}(P,10),\; O_A^{inq} \bigr)
\end{split}
\end{equation}

We then select every hard example plus 20\% as many easy ones sampled round-robin across integer difficulty bins to maintain a balanced curriculum.

The second example, designated as a “difficulty‐prediction” instance, teaches Alice to self‐assess the hardness of its own creations: by taking Alice's generated program $Q$ along with the Difficulty Prediction User Prompt and supervising on the numeric label ``$\text{Difficulty level:} \hat d $" (Schema~\ref{Alice_second_example}). The training dataset is also biased towards hard examples, with easy examples comprising 50\% of the subset. Alice learns to calibrate its difficulty estimates and ensures that its future generations are appropriately challenging. 
\begin{equation}\label{Alice_second_example}
\begin{split}
    \text{SEQ}_{\text{DP}} &= \bigl(\mathrm{SP}_{A,DP}^{eq},\; \mathrm{UP}_{A,DP}^{eq},\; \hat{d} \bigr) \\
    \text{SINQ}_{\text{DP}} &= \bigl(\mathrm{SP}_{A,DP}^{inq},\; \mathrm{UP}_{A,DP}^{inq},\; \hat{d} \bigr)
\end{split}
\end{equation}

The third example concerns all $(P, Q)$ pairs in the SEQ that have been successfully proved. We not only consider the generated candidate $Q$ for a given $P$, but also the complete reasoning trace and valid Liquid Haskell proof script from Alice (Schema~\ref{Alice_third_example}). These examples serve as supervised training signals to help Alice internalize the proof obligation $\pi$, corresponding to the refinement type $\forall x., P(x) = Q(x)$.
\begin{equation}\label{Alice_third_example}
\begin{split}
    \text{SEQ}_{\text{proof}} = \Big\{ \bigl(&\mathrm{SP}_{A}^{eq},\; \mathrm{UP}_{A}^{eq}(P),\; O_{A}^{eq}(P,Q,\pi)\bigr) \\
    &\;\Big|\; \mathrm{LH}(P,Q)\vdash \pi \Big\}
\end{split}
\end{equation}

This three-part structure enables Alice to generate challenging variants, assess their difficulties, and internalize proof strategies.

\subsubsection{Bob's Training Data Selection}

Bob’s training data is constructed from all of his correct $(P, Q)$ pairs' evaluations. Each training example (Schema~\ref{Bob_example}) consists of the original pair, the system prompts, user prompts, and Bob’s response $O_B$. By sampling from a wide range of difficulties, Bob learns to reason and recognize both easy and subtle equivalence relations.
\begin{equation}\label{Bob_example}
    \mathcal{E}_{\text{Bob}} = \bigl(\mathrm{SP}_B,\; \mathrm{UP}_B(P,Q),\; O_B \bigr)
\end{equation}

\section{Experimental Setup}\label{sec:main_experiment_setup}

We use \texttt{DeepSeek-R1-Distill-Qwen-7B} as our base model for both Alice and Bob. Fine-tuning is performed using LoRA adaptors. Please see Appendix~\ref{app:main_exp_setup} for more details.

To investigate the distinct contributions of equivalence versus inequivalence supervision, we conduct the main experiment ($E_0$) alongside three controlled ablations ($E_1$--$E_3$). These regimes systematically vary the game type probability and reference program budget ($P$) to isolate the impact of SEQ supervision under controlled data constraints. The configurations are shown in Table~\ref{tab:regimes}.

\begin{table}[H]
\centering
\small
\resizebox{\columnwidth}{!}{%
\begin{tabular}{l|ccc}
\toprule
\textbf{Regime} & \textbf{SEQ/SINQ} & \textbf{$P$} & \textbf{Goal} \\
\midrule
$E_0$ (Base) & 50/50 & 500 & Main \\
$E_1$ & 0/100 & 500 & Max Volume \\
$E_2$ & 96/4 & 500 & Balanced Yield \\
$E_3$ & 0/100 & 40 & Vol. Control ($E_3 \approx E_2$) \\
\bottomrule
\end{tabular}%
}
\caption{Experimental regimes to test SEQ impact.}
\label{tab:regimes}
\end{table}

The full codebase is available at the GitHub repository\footnote{\url{https://github.com/TrevorPoon/llm-self-play-liquidhaskell}} to generate the dataset OpInstruct-HSx, reproduce the experiments and evaluate model performance.

\section{Results}

The following evaluation results are structured to address the three research questions outlined in the introduction. 

\subsection{Intrinsic Evaluation}

The goal of $E_0$ is teaching Alice (the generator) to generate increasingly challenging instances while Bob (the evaluator) to improve at judging program semantics.

\subsubsection{Capability Assessment}

\begin{figure}[htbp]
    \centering
    \includegraphics[width=0.85\linewidth]{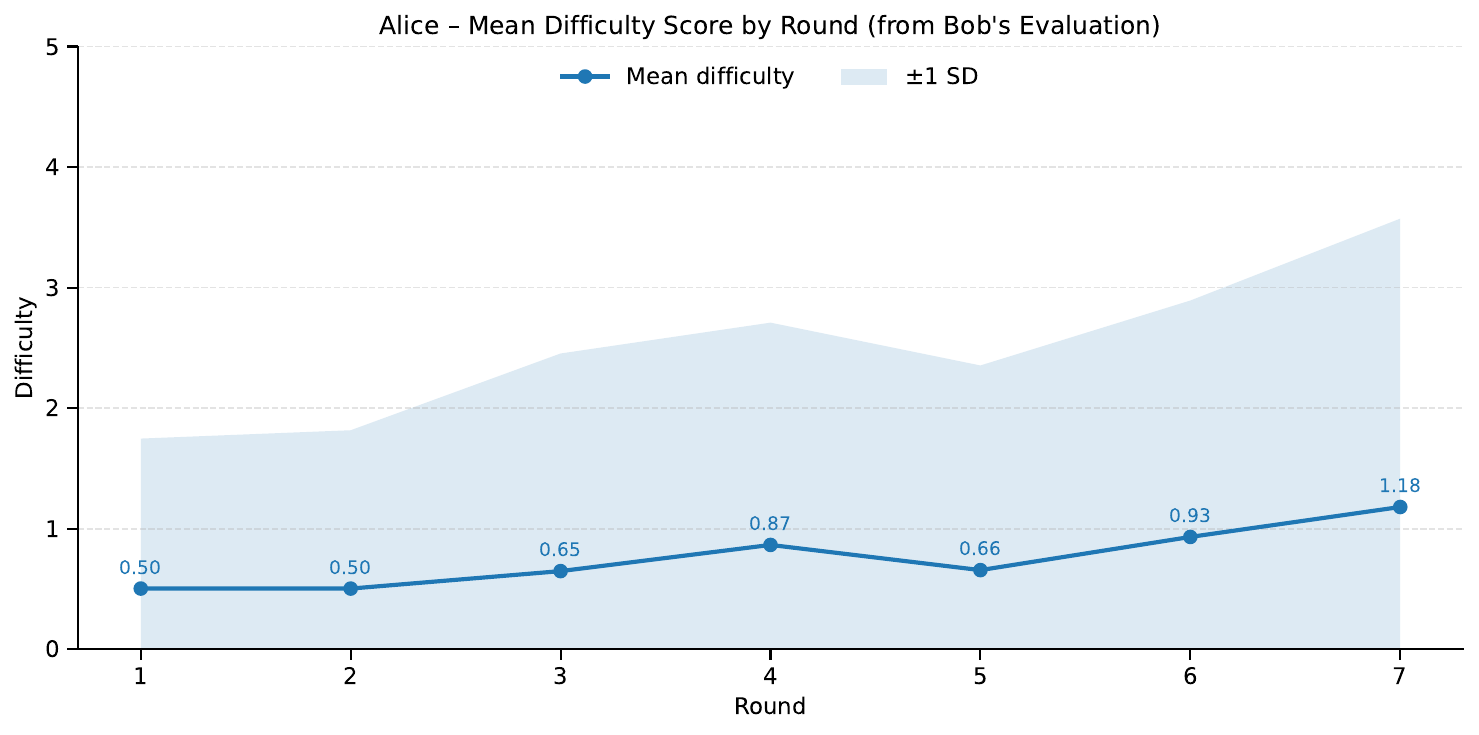}
    \caption{Mean and standard deviation for the difficulty scores for Alice's generated instances from a fixed untrained Bob over 7 rounds.}
    \label{fig:diff_chart}
\end{figure}

As shown in Figure~\ref{fig:diff_chart}, the mean difficulty rises from 0.50 to 1.18 by round 7, indicating that Alice is crafting harder instances for a constant Bob. See Appendix~\ref{app:difficulty_trajectory} for the difficulty trajectories breakdown for both SEQ and SINQ games.

In addition, we measure how much the evaluator model (Bob) improves after its first and only training round in the semantic equivalence self-play. Challenge instances are generated by the final trained generator model, Alice from round 7, using source programs from the training split of the OpInstruct-HSx and from the unseen test split. The resulting accuracies, summarised in Table \ref{tab:opinstruct_hsx_accuracy}, show modest improvements on unseen data.

\newcolumntype{Y}{>{\centering\arraybackslash}X}
\newcolumntype{L}{>{\raggedright\arraybackslash}X}
\newcommand{\imp}[1]{\textcolor{green!50!black}{#1}}   % improvement
\newcommand{\drop}[1]{\textcolor{red!50!black}{#1}}    % drop is good for errors

\begin{table}[htbp]
\centering
\small 
\begin{tabularx}{\linewidth}{
    >{\raggedright\arraybackslash}p{3cm}  
    >{\raggedright\arraybackslash}p{1cm}  
    >{\centering\arraybackslash}p{1cm}    
    >{\centering\arraybackslash}p{1cm}  
    }
\toprule
& \multicolumn{3}{c}{\textbf{Accuracy (\%)}} \\
\cmidrule(lr){2-4}
\textbf{Benchmark} & \textbf{Base} & \textbf{Trained} & \textbf{$\Delta$} \\
\midrule
OpInstruct-HSx (Train) & 87.57 & \textbf{91.34} & \imp{$+3.77$} \\
OpInstruct-HSx (Test)  & 88.24 & \textbf{88.79} & \imp{$+0.55$} \\
\bottomrule
\end{tabularx}
\caption{OpInstruct-HSx accuracy results comparing the Base Model and Trained (Bob adapter) models.}
\label{tab:opinstruct_hsx_accuracy}
\end{table}

\subsection{Extrinsic Evaluation}

Having validated the agents' improvement within the self-play loop, we now investigate whether this training enhances the model's understanding of program semantics across different programming languages and domains.

\subsubsection{Haskell Program Generation}

In MBPP and HumanEval from MultiPL-E~\citep{cassano2022multiplescalableextensibleapproach}, both models' compilation errors decrease substantially and Pass@1 improves on the Haskell tasks (shown in Table~\ref{tab:he_mbpp_results}).  Performance trajectories for Alice across self-play rounds are provided in Appendix~\ref{app:e0_alice_Humaneval} and ~\ref{app:e0_alice_MBPP}.

\begin{table*}[t]
\centering
\small
\begin{tabularx}{\textwidth}{
    >{\raggedright\arraybackslash}p{2.5cm}  
    >{\raggedright\arraybackslash}p{1.5cm}  
    Y  
    Y  
    Y  
    Y  
    Y  
    Y  
}
\toprule
& & \multicolumn{3}{c}{\textbf{Pass@1 (\%)}} 
& \multicolumn{3}{c}{\textbf{Compilation Errors}} \\
\cmidrule(lr){3-5}\cmidrule(lr){6-8}
\textbf{Benchmark} & \textbf{Agent}
& \textbf{Base} & \textbf{Trained} & \textbf{$\Delta$}
& \textbf{Base} & \textbf{Trained} & \textbf{$\Delta$} \\
\midrule
HumanEval & Bob & 17.7 & \textbf{26.4} & \imp{$+8.7$}
          & 130  & \textbf{110}  & \imp{$-20$} \\
HumanEval & Alice & 17.7 & \textbf{26.3} & \imp{$+8.6$}
          & 130  & \textbf{104}  & \imp{$-26$} \\
MBPP      & Bob   & 26.7 & \textbf{36.9} & \imp{$+10.2$}
          & 250  & \textbf{203}  & \imp{$-47$} \\
MBPP      & Alice   & 26.7 & \textbf{34.3} & \imp{$+7.6$}
          & 250  & \textbf{218}  & \imp{$-32$} \\
\bottomrule
\end{tabularx}
\caption{HumanEval and MBPP results (zero-shot prompting in Haskell), comparing the Base Model (untrained) and Trained models. Averages over 16 trials.}
\label{tab:he_mbpp_results}
\end{table*}

The results indicate that semantic reasoning training transfers positively to Haskell code generation performance and robustness for both agents. For Bob, the gains confirm that the semantic discrimination skills learned during self-play can yield broader benefits in practical programming scenarios. And for Alice, the improvements suggest that training on high-quality, verifiable program transformations in self-play can enhance downstream synthesis accuracy and reduce compilation errors.

\subsection{Coding Related Tasks Evaluation}

We would like to investigate whether richer type semantics learnt in the evaluator model Bob can further enhance his ability to analyse correct and semantically meaningful code across other coding paradigms, not just the functional ones.

\subsubsection{EquiBench}

The EquiBench~\citep{wei2025equibench} benchmark consists of six data categories: DCE (C pairs with dead/live code variations), x86-64 (identical assembly sequences via superoptimization), and CUDA (equivalent kernels with different scheduling) form the low-level systems group. The algorithmic group uses Python pairs from competitive programming, featuring OJ\_A (algorithmic refactors), OJ\_V (variable renaming), and OJ\_VA (combined algorithmic and variable changes). 

In the zero-shot prompting evaluation, our fine-tuned Bob model shows clear gains over the base model in several dataset configurations.

\begin{figure}[htbp]
    \centering
    \includegraphics[width=0.95\linewidth]{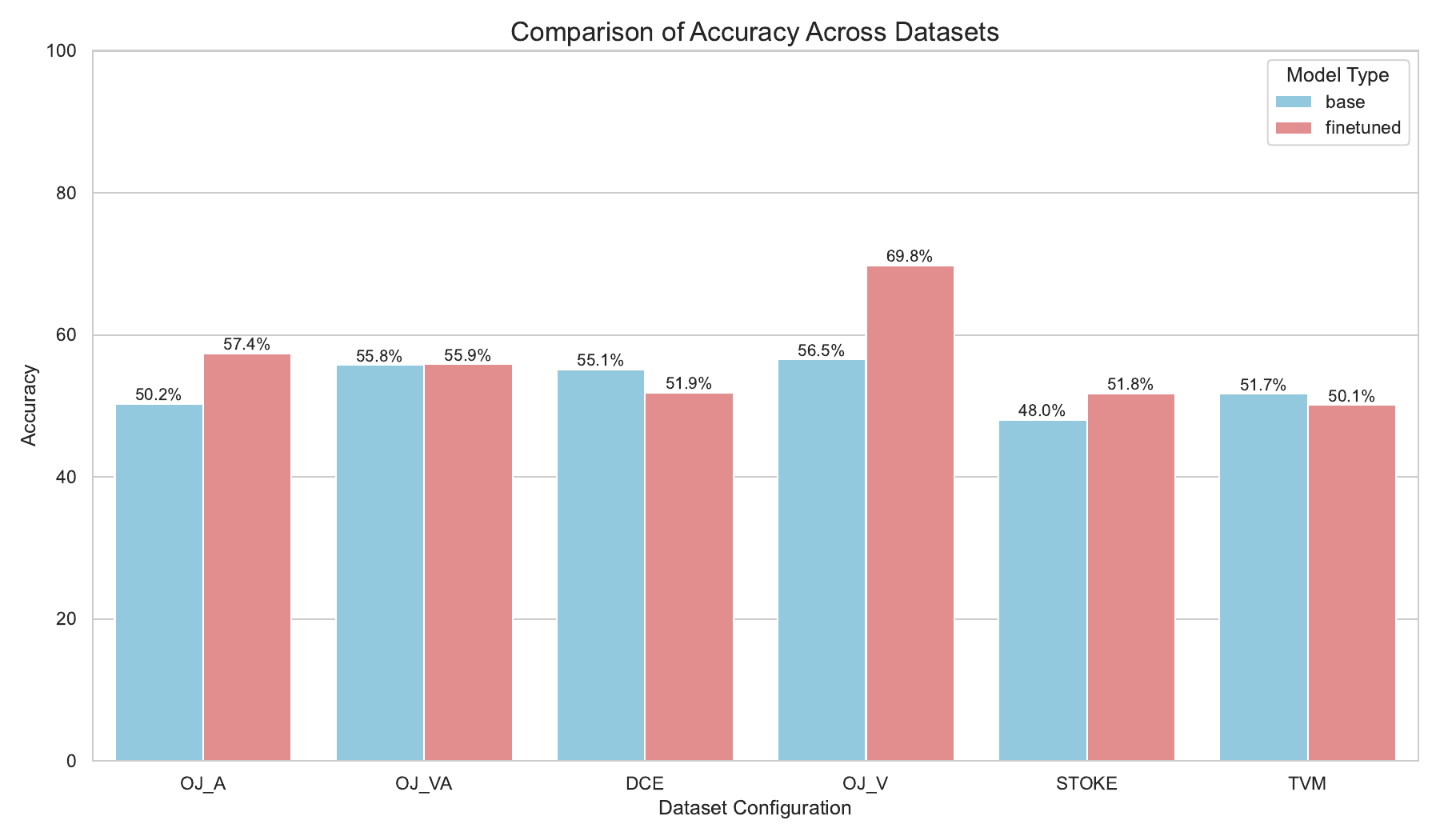}
    \caption{Accuracy comparison between base and fine-tuned Bob models across EquiBench dataset configurations.}
    \label{fig:accuracy_comparison_EquiBench}
\end{figure}

\begin{figure}[htbp]
    \centering
    \includegraphics[width=0.95\linewidth]{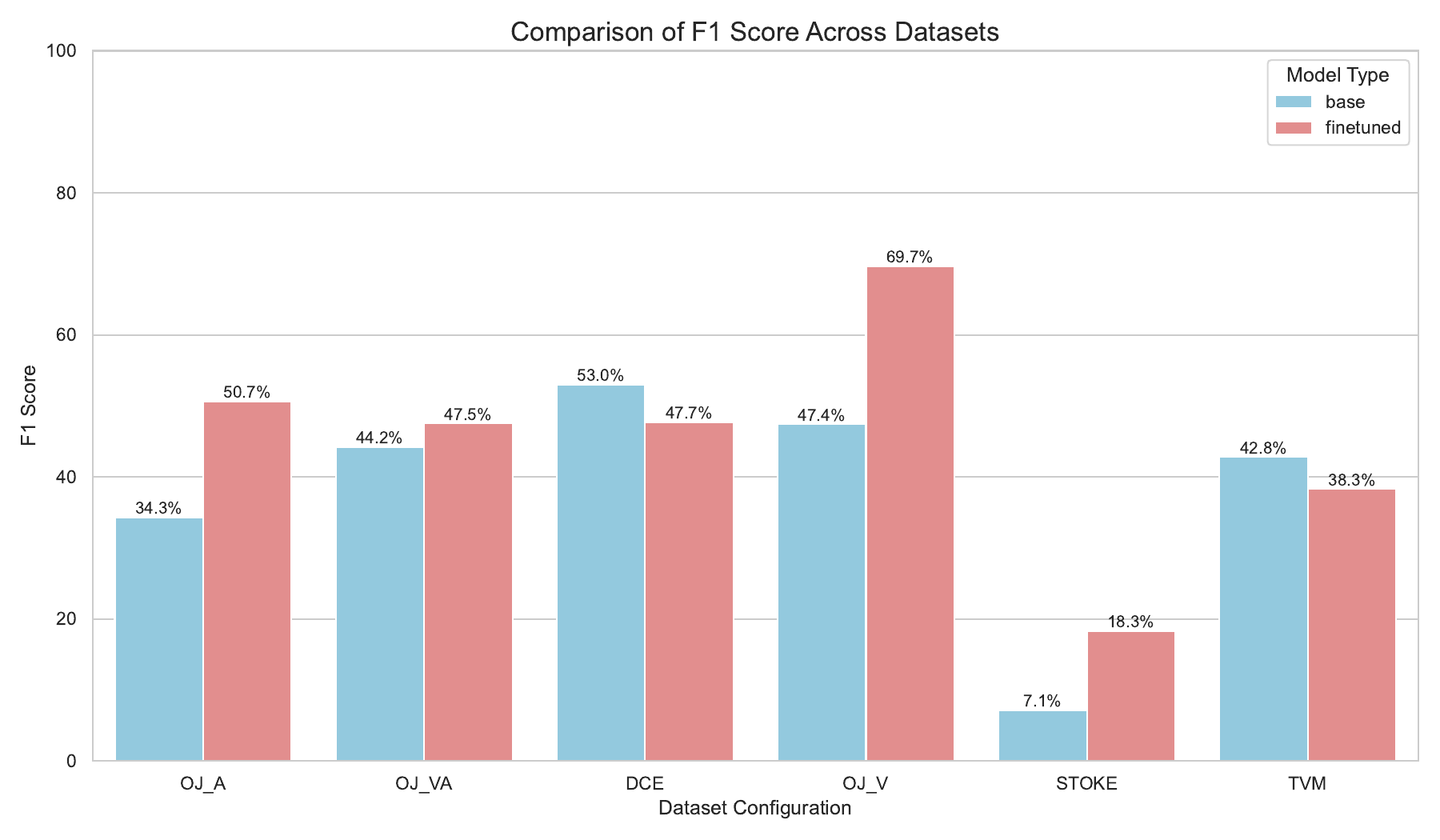}
    \caption{F1 score comparison between base and fine-tuned Bob models across EquiBench dataset configurations.}
    \label{fig:f1_score_comparison_EquiBench}
\end{figure}

From Figure~\ref{fig:accuracy_comparison_EquiBench} and Figure~\ref{fig:f1_score_comparison_EquiBench}, the model transfers best to OJ\_A and OJ\_V because both categories (algorithmic refactors and variable renaming respectively) align with Bob’s learned skill on detecting high-level behavioural divergence, resulting in large gains in performance.  OJ\_VA's performance stays modest as the authors stated that it is harder than pure renaming and closer to the “non-local structural” end.

In contrast, DCE, STOKE, and TVM require reasoning about low-level or non-functional semantics absent from our Haskell curriculum. This mismatch in language, abstraction level, and equivalence definition explains the weaker or negligible improvements on these tasks.

\subsubsection{PySecDB}

PySecDB is the first comprehensive dataset of security-related commits in Python~\citep{sun2023exploringsecuritycommitspython}. In Table~\ref{tab:pysecdb_results}, our fine‐tuned Bob model achieves consistent improvements over the base model across all evaluated metrics. The results indicate that Bob’s Haskell semantic reasoning training transfers positively to Python vulnerability detection, improving in identification of security‐relevant code changes.

\begin{table}[htbp]
\centering
\small
\begin{tabularx}{\linewidth}{LYYYYY}
\toprule
& \multicolumn{3}{c}{\textbf{Scores (\%)}} \\
\cmidrule(lr){2-4}
\textbf{Metric} 
& \textbf{Base Model} & \textbf{Trained} & \textbf{$\Delta$} \\
\midrule
Accuracy & 67.6 & \textbf{68.8} & \imp{$+1.2$} \\
Precision & 48.0 & \textbf{49.7} & \imp{$+1.7$} \\
Recall    & 55.1 & \textbf{59.2} & \imp{$+4.1$} \\
F1 Score  & 51.3 & \textbf{54.0} & \imp{$+2.7$} \\
\bottomrule
\end{tabularx}
\caption{PySecDB vulnerability detection results comparing the Base Model and Trained (Bob adapter).}
\label{tab:pysecdb_results}
\end{table}

\subsubsection{CodeXGlue}

CodeXGLUE’s defect detection dataset~\citep{zhou2019devign} is a curated collection of over 27,000 C-language functions, each manually labeled to indicate whether it contains a security-relevant defect.

\begin{table}[htbp]
\centering
\small
\begin{tabularx}{\linewidth}{LYYYYY}
\toprule
& \multicolumn{3}{c}{\textbf{Scores (\%)}} \\
\cmidrule(lr){2-4}
\textbf{Metric} 
& \textbf{Base Model} & \textbf{Trained} & \textbf{$\Delta$} \\
\midrule
Accuracy & 48.8 & \textbf{50.5} & \imp{$+1.7$} \\
Precision & 45.0 & \textbf{45.7} & \imp{$+0.7$} \\
Recall    & \textbf{51.1} & 41.7 & \drop{$-9.4$} \\
F1 Score  & \textbf{47.9} & 43.2 & \drop{$-4.7$} \\
\bottomrule
\end{tabularx}
\caption{CodeXGLUE defect detection results comparing the Base Model and Trained (Bob adapter).}
\label{tab:codexglue_results}
\end{table}

In Table~\ref{tab:codexglue_results},  both the base and fine‐tuned Bob models perform near the 50\% mark for accuracy, which is indistinguishable from a random model. CodeXGLUE’s dataset gives Bob only one C function and asks if it’s vulnerable. The vulnerabilities are low-level memory issues that require tracking pointers, memory allocation, and data layouts. However, Haskell is memory-safe, garbage-collected, and doesn’t have manual frees or pointer arithmetic. Therefore, the trained model shows no meaningful advantage over the base model on this task. 

\subsection{Controlled Comparison of SEQ--SINQ Regimes}\label{sec:controlled_comparison_of_regimes}

\begin{figure}[htbp] % or [htbp]
    \centering
    \includegraphics[width=\linewidth]{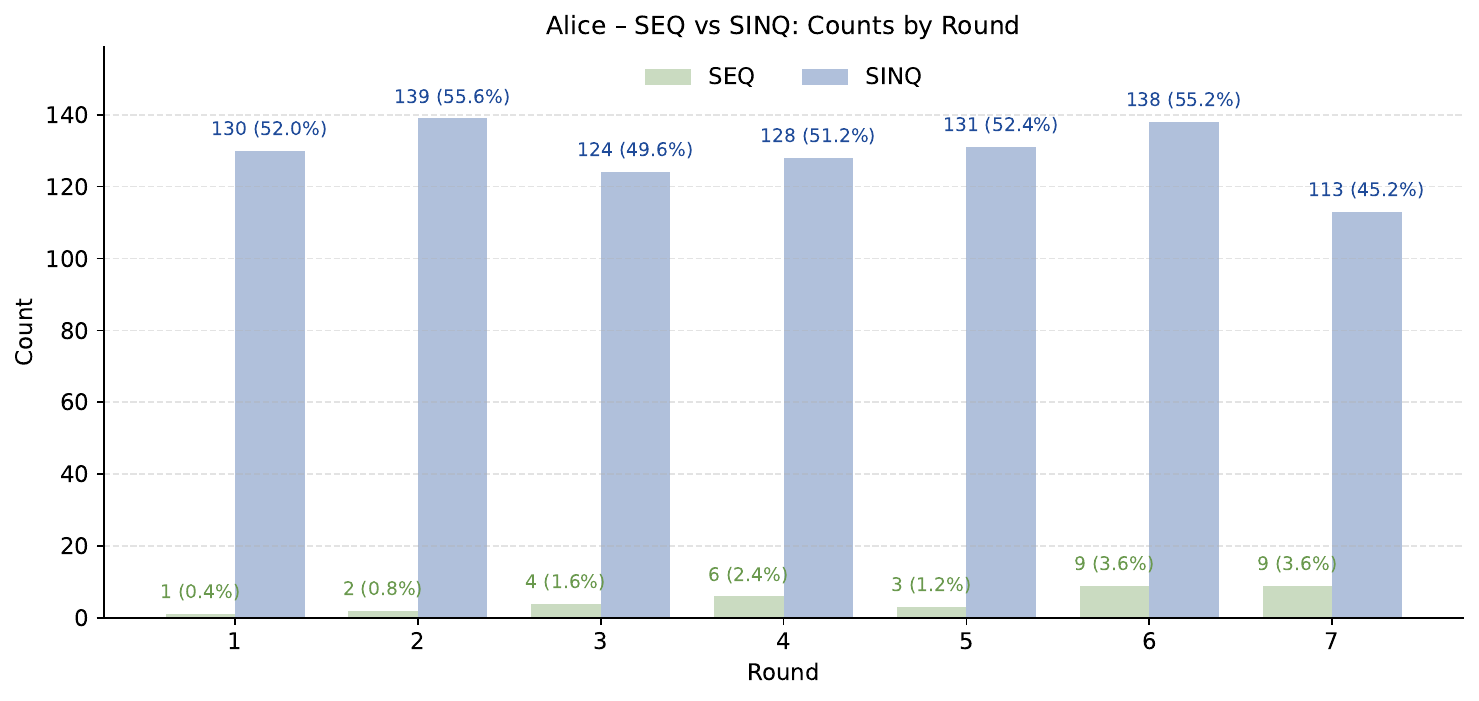}
    \caption{Counts and proportions of validated Alice generations by round. }
    \label{fig:count_by_round}
\end{figure}

A critical finding from $E_0$ is the asymmetry in number of validated programs between SINQ and SEQ even under a uniform sampling policy in choosing the game. In Figure~\ref{fig:count_by_round}, we have hundreds of accepted SINQ instances per round, but very few SEQ instances (single digits in most rounds). A review of the output from the self-play shows that while Alice can readily produce SINQ pairs, the small reasoning model often struggles to produce Liquid Haskell proofs for SEQ, which drastically limits the number of compiled and verified SEQ examples available for training. This imbalance limits the diversity of program types in the training buffer, with the current SEQ+SINQ configuration heavily skewed with SINQ examples.

This subsection investigates whether incorporating SEQ supervision yields unique reasoning benefits beyond those of SINQ alone, aiming to isolate the impact of the supervision type from the disparity in verified data volume (see Appendix~\ref{app:com_exp} for derivation details).

\subsubsection{Evaluation Results}

We analyze the four experimental regimes to decouple the effects of supervision type (SEQ vs. SINQ) from training data volume. All results are presented in Appendix~\ref{app:exp_result} and~\ref{app:eval_result}.

\subsubsection*{Value of SEQ Supervision}
Despite having significantly fewer verified training pairs in $E_0$ than in the pure SINQ regime $E_1$ ($937$ vs. $1,839$), $E_0$ achieves superior performance on semantic equivalence tasks. This indicates that SEQ supervision improves high-level semantic judgment even when data volume is reduced. Furthermore, when strictly controlling for data volume ($E_2$ vs. $E_3$, both $\approx 150$ pairs), the inclusion of SEQ ($E_2$) yields consistent advantages on structural reasoning tasks over pure SINQ ($E_3$), shown in Table~\ref{tab:benchmark_perf_all_exp}. This confirms that SEQ supervision confers a unique benefit that cannot be replicated by inequivalence signals alone.

\subsubsection*{Volume Trade-offs}
While SEQ is qualitatively valuable, data volume remains critical. Regime $E_2$, which aggressively prioritizes SEQ (96\% attempts) to force a balanced validated distribution, suffers a sharp drop in total verified pairs ($N=140$) due to low proof yields, causing performance degradation across all benchmarks compared to $E_0$. The 50/50 mixed regime ($E_0$) therefore represents an acceptable trade-off, balancing the semantic depth of SEQ with the high verification yield of SINQ.

\section{Conclusions}

This paper investigates the use of self-play to improve code generation and semantic reasoning in LLMs, focusing on program equivalence in Haskell and Liquid Haskell. The framework introduces two agents: a generator of semantically equivalent or inequivalent program variants, and an evaluator trained to judge equivalence. Evaluation shows that Bob benefits substantially: Haskell coding performance improves, and transfer is strong on high-level semantic reasoning tasks, but negligible on low-level or memory-oriented tasks. Controlled comparisons reveal that although inequivalence data yields far more verified pairs, the inclusion of equivalence supervision confers unique gains in semantic reasoning that cannot be achieved through inequivalence alone.

\section{Future Work}

There are several promising avenues for future research. First, extending training across more rounds could allow Bob to accumulate richer experience. Additionally, (1) scaling to larger parameter models and (2) employing full-parameter fine-tuning instead of LoRA may increase both Alice’s proof synthesis capabilities and Bob’s reasoning capacity.

A second priority is to improve the statistical robustness of the controlled comparisons. The experiments in regimes $E_2$ and $E_3$ were limited by small sample sizes, which may not be representative of the true differences between SINQ- and SEQ-based supervision. Running larger-scale experiments would provide stronger evidence about the relative contributions of the two regimes.

Furthermore, methodologically speaking, the Haskell self-play framework could be expanded to cover low-level program behaviors by introducing tasks that explicitly model memory usage, side effects, and bit-level operations. 

Moreover, converting the self-play pipeline into a dedicated Haskell Equivalence Evaluation test set would help to provide a more direct and sustainable benchmark for semantic reasoning, especially given the scarcity of high-quality Haskell resources currently available.

Finally, reinforcement learning could be re-examined as a means of enhancing the SEQ game’s contribution. By running Alice long enough to generate sufficiently challenging programs and providing Bob with richer feedback through reward-based updates, it may be possible to overcome the current proof bottleneck and more effectively integrate equivalence reasoning into the self-play loop.

\section{Limitations}

A central limitation of this work lies in the proof synthesis bottleneck. Alice rarely produces Liquid Haskell proofs that successfully pass PLE, resulting in very low validation yields for SEQ cases. This imbalance causes the majority of verified training data SINQ-dominated, reducing the contribution of SEQ supervision to Bob’s learning. The issue is likely exacerbated by the relatively small model size used in this study, which likely restricts Alice’s ability to generate structurally complex proofs.

Another limitation arises from the inherent constraints of Liquid Haskell itself. The verification process depends heavily on reflection and proof by logical evaluation, which are effective for local reasoning but struggle with certain classes of programs. Non-terminating behaviors, partial functions, and large-scale algebraic rewrites are difficult to certify, and in some cases impossible, within this fragment of the logic. Moreover, not all Haskell programs are reflectable, further narrowing the scope of tasks where equivalence can be formally verified.

Finally, there is a cross-domain mismatch in generalization. While Bob transfers strongly to high-level semantic reasoning tasks, such as OJ\_A and OJ\_V in EquiBench and vulnerability detection in PySecDB, his performance is notably weaker on low-level or stateful semantics, including DCE, STOKE, TVM, and CodeXGlue. This gap highlights a limitation of the current framework in addressing equivalence reasoning that depends on memory, side effects, or bit-level operations.

\section{Ethical considerations}

This work adversarially trains models to introduce hard to detect semantic modifications to computer programs and then to detect them, which may train them to detect security-relevant vulnerabilities or introduce obfuscated backdoors.
In principle, these capabilities have a potential for malicious use, however, they can be also used for pro-social use to improve the reliability and security of code, whether generated by humans or LLMs.
We believe that the net societal impact of models better able to reason about the subtleties of program semantics to be positive, especially as these capabilities are disseminated through Open Source releases, as we do in this work.

\section*{Acknowledgments}

This work has been supported by the UK government though the AISI-EPSRC Systemic Safety Grant titled "\textbf{Understanding and Improving the Behaviour of AI Agents in Competitive and Cooperative Games}" and by Amazon through the AWS Agentic Amazon Research Award titled "\textbf{Diffusion-inspired chain-of-thought self-revision}".
The authors also thank Shay Cohen and Vaishak Belle for their advice.

\bibliography{custom}

\clearpage

\appendix

\section{Illustrative Haskell Features for Semantic Reasoning and Verification in LLM Self‐Play}\label{app:haskell_features}

\lstset{basicstyle=\ttfamily\scriptsize,breaklines=true,columns=fullflexible}

\tcbset{
  myboxstyle/.style={
    enhanced,
    colback=blue!5,
    colframe=black,
    fonttitle=\bfseries,
    boxsep=4pt,
    left=2pt, right=2pt, top=2pt, bottom=2pt,
    halign title=flush left,
    % Removed 'equal height group' so boxes adapt to their content height
  }
}

\lstdefinestyle{smallcode}{
  basicstyle=\ttfamily\tiny,
  breaklines=true,
  columns=fullflexible,
  keepspaces=true,
  showstringspaces=false,
}

\begin{figure}[h]
  \centering
  \scriptsize
  
  % --- Top Box ---
  \begin{tcolorbox}[
      title={[1] Pure Semantics},
      myboxstyle
  ]
    Every function is a pure mapping from inputs to outputs, there is no hidden state or mutation. This purity ensures that two functions are semantically equivalent if they produce the same outputs for all inputs, regardless of how those outputs are computed. \\[2pt]
    \begin{lstlisting}[style=smallcode]
-- revRec :: [a] -> [a]
revRec :: [a] -> [a]
revRec []     = []
revRec (x:xs) = revRec xs ++ [x]
-- revFold :: [a] -> [a]
revFold :: [a] -> [a]
revFold = foldl (flip (:)) []
    \end{lstlisting}
    Because Haskell is referentially transparent, \texttt{revRec xs} can be substituted with \texttt{revFold xs} wherever it appears. This makes reasoning about equivalence both feasible and formalizable.
  \end{tcolorbox}

  % --- Vertical Spacing ---
  \vspace{0.2cm} 

  % --- Bottom Box ---
  \begin{tcolorbox}[
      title={[2] Static Typing and GHC Compile},
      myboxstyle
  ]
    Haskell's static type system provides strong guarantees at compile time. Once a program is accepted by the compiler, most classes of errors such as type mismatches and null de-referencing are eliminated. This makes the type checker an effective pre-filter for program validity in the self-play loop.\\[2pt]
    \begin{lstlisting}[style=smallcode]
-- add :: Int -> Int -> Int
add x y = x + y
    \end{lstlisting}
    When writing \texttt{add True False}, GHC will raise a compilation error before running the program as \texttt{add} expects two \texttt{Int}. 
  \end{tcolorbox}
  \label{fig:haskell-features}
\end{figure}

\section{Liquid Haskell SEQ / SINQ Instances}\label{app:LH_instances}

\begin{lstlisting}[caption={Example of a SEQ Instance},numbers=none]
-- Reference Function P
double :: Int -> Int
double x = x + x

-- Alice's generated Q (Semantic Equivalent)
double_alt :: Int -> Int
double_alt x = 2 * x

-- Liquid Haskell Proof
{-@ lemma_double_equiv :: x:Int 
      -> { double x == double_alt x } @-}
lemma_double_equiv :: Int -> Proof
lemma_double_equiv x
  =   double x
  === double_alt x
  *** QED
\end{lstlisting}

\begin{lstlisting}[caption={Example of a SINQ Instance},numbers=none]
-- Original function P
sign :: Int -> String
sign n
  | n < 0     = "negative"
  | n == 0    = "zero"
  | otherwise = "positive"

-- Alice's generated Q (semantically inequivalent)
signIneq :: Int -> String
signIneq n
  | n <= 0    = "non-positive"
  | otherwise = "positive"

-- Diverging input
x_a = 0
-- P x_a = sign 0 = "zero"
-- Q x_a = signIneq 0 = "non-positive"
\end{lstlisting}

\section{Implementation with Reinforcement Learning}\label{app:RL}

Reinforcement learning (RL) offers a conceptually natural way to optimize the game by directly rewarding or penalizing generated programs based on the game outcome. A straightforward RL setup would proceed as follows based on the game rules:

\subsection{RL Formulation}

We treat Alice and Bob as two competing agents in a zero-sum Markov game. 
\begin{itemize}
    \item For the SINQ branch: An executor first checks that $Q$ compiles and that $P(x_a)\neq Q(x_a)$ on Alice proposed diverging input $x_a$.
    \item For the SEQ branch: the SMT‐based checker will execute the Liquid Haskell proof script asserting $ \forall x.\;P(x)=Q(x) $.
\end{itemize}

If the above verification fails, the episode terminates with Alice receiving a negative reward $r_A=r_{\mathrm{fail}}<0$. Otherwise Bob observes state $(P,Q)$ (but not Alice’s diverging input $x_a$) and identify whether $(P,Q)$ are semantically equivalent.  

If Bob is correct, then Bob earns $r_B=r_{\mathrm{success}}>0$ and Alice is penalized with $r_A=r_{\mathrm{too\_easy}}<0$.  If Bob is incorrect, he gets $r_B=r_{\mathrm{fail}}<0$ and Alice wins $r_A=r_{\mathrm{win}}>0$.  Both agents update their policies to maximize expected cumulative reward over many self‐play episodes.

\subsection{Potential Benefits of RL}

RL can be layered on top of the semantic equivalence self‐play by treating Alice’s generator as an RL agent whose policy is updated based on proof outcomes. Over many episodes, this encourages the policy to avoid classes of common semantic errors (e.g. off‐by‐one edge cases) and refines its internal value function to distinguish subtle equivalence‐breaking patterns, resulting in an LLM that is both more precise and robust.

\subsection{Practical Challenges}

\begin{itemize}
   \item \textbf{Sparse, High‐Variance Rewards}: In our reasoning tasks, reward is only provided at the end of a multi-step chain of thought. As a result, the LLMs often struggle to find successful trajectories, making it difficult for LLMs to learn which intermediate steps contributed to success~\citep{liu2025superrlreinforcementlearningsupervision}.
   \item \textbf{Credit Assignment}: Attributing success or failure back to specific reasoning tokens or code‐editing steps is nontrivial, and naive reward assignment can lead to undesirable shortcuts, such as inserting semantically neutral modifications to exploit reward signals without genuine reasoning.
  \item \textbf{Computational Resources Constraints}: RL is more computationally expensive than supervised fine-tuning due to the repeated sampling, model rollouts, and reward computation across multiple steps~\citep{ouyang2022traininglanguagemodelsfollow}.
\end{itemize}

\lstdefinelanguage{nocode}{} 
\lstdefinestyle{promptstyle}{
    language=nocode,                 % <-- disable keyword set
    basicstyle=\ttfamily\footnotesize\mdseries,
    keywordstyle=\ttfamily\footnotesize\mdseries, % <-- override bold
    commentstyle=\ttfamily\footnotesize\mdseries,
    stringstyle=\ttfamily\footnotesize\mdseries,
    emph={}, emphstyle=\ttfamily\footnotesize\mdseries, % <-- no emph
    morekeywords={},                 % <-- clear any inherited keywords
    numbers=none,                    % <-- no line numbers
    breaklines=true,
    breakatwhitespace=true,
    frame=single,
    backgroundcolor=\color{gray!10},
    columns=fullflexible,
    keepspaces=true,
    aboveskip=2pt, belowskip=2pt,
    lineskip=-1pt,
    xleftmargin=1pt, xrightmargin=1pt,
    framexleftmargin=1pt, framexrightmargin=1pt,
    framextopmargin=1pt, framexbottommargin=1pt,
    showstringspaces=false,
    upquote=true
}
\lstset{style=promptstyle}

\section{Prompt Template}\label{both_prompt}

\subsection{Alice SEQ System Prompt}

\begin{lstlisting}[mathescape=true]
You are a helpful and expert Haskell programmer, powered by Liquid Haskell.
Your job is to *transform* any given function `P` into a new function `Q` that:

  - Is syntactically correct Haskell.  
  - Is semantically equivalent: $\forall$x. `P x == Q x`.  
  - Uses a *different* implementation -- Implement a non-structural change rather than merely swapping operator order. 
  - Uses a different function name (e.g. add a trailing `'_alt`).  

The original program and your program will be used in a test to evaluate the skill of an expert Haskell programmer who will identify whether your program Q is semantic equivalent to the original program P, so make sure that the difference you introduce are not very easy to understand. 
You will be given a difficulty level from 0 (easiest) to 10 (hardest) to target. E.g. difficulty level 0 means that an expert computer scientist in the bottom decile or above should be able to find a diverging example, difficulty level 9 means that only an expert computer scientist in the top decile should be able to find a diverging example, and difficulty level 10 means that only the top 0.01 or less of expert Haskell programmer should be able to find a diverging example.                                 

Always think through your transformation steps in `<think>...</think>`, then emit exactly:

Generated Program `Q`:
```haskell
<your Q here>
```
\end{lstlisting}

\subsection{Alice SEQ User Prompt}\label{app:Alice_SEQ_User_Prompt}

\begin{lstlisting}
Difficulty level: {difficulty_level}
Here is the original Haskell function `P`:
```haskell
{program_p_completion}
```

Its argument type is  
```haskell
t = {t}
```

Your task: produce a new function `Q` that satisfies the system prompt requirements.  
- Make sure `Q` has a different name (e.g. append a `'_alt`).  
- Avoid trivial symmetric rewrites - show a genuine alternative implementation.  
- Do not include any extra commentary beyond the required `<think>...</think>` and the `Generated Program `Q`: block.
- Where appropriate, feel free to use Prelude functions such as foldr, map, or zipWith to encourage diverse strategies.

<think>
\end{lstlisting}

\subsection{Lemma SEQ Proof System Prompt}
\begin{lstlisting}
You are an expert Haskell/Liquid Haskell prover.
You are asked to prove that two reflected functions are equivalent.
                                        
The most basic proof should be in the following format:
```haskell
{{-@ lemma_{func_name_p}_equiv :: x:{arg_type} -> {{ {func_name_p} x == {func_name_q} x }} @-}}
        lemma_{func_name_p}_equiv :: {arg_type} -> Proof
        lemma_{func_name_p}_equiv x
            =   {func_name_p} x 
            === {func_name_q} x 
            * QED
```
                                        
However, you should also use more advanced proof techniques if necessary. 
                                        
Few-Shot Example 1:

```haskell
{{-@ LIQUID "--reflection" @-}}
{{-@ LIQUID "--ple" @-}}

module MyTest where

import Language.Haskell.Liquid.ProofCombinators

-- Alice program P
{{-@ reflect double @-}}
double :: Int -> Int
double x = x + x

-- Alice proposes Q
{{-@ reflect double' @-}}
double' :: Int -> Int
double' x = 2 * x

-- Here is the full lemma, from annotation to QED:
{{-@ lemma_double_equiv :: x:Int -> {{ double x == double' x }} @- }}
lemma_double_equiv :: Int -> Proof
lemma_double_equiv x
=   double x
=== double' x
* QED
```
                                        
Few-Shot Example 2:
```haskell
{{-@ LIQUID "--reflection" @-}}
{{-@ LIQUID "--ple" @-}}
module Equiv where

import Language.Haskell.Liquid.ProofCombinators

{{-@ reflect addNumbers @-}}
addNumbers :: Int -> Int -> Int
addNumbers a b = a + b

{{-@ reflect addNumbers' @-}}
addNumbers' :: Int -> (Int -> Int)
addNumbers' a = \b -> a + b

-- Alice detailed proof of equivalence
lemma_addNumbers_equiv :: Int -> Int -> Proof
lemma_addNumbers_equiv x y
    =   addNumbers x y 
    === addNumbers' x y 
    * QED
                                        
When you answer, output only the complete lemma block in the same style:
1. Use the `{{-@ lemma_... @-}}` annotation , with the exact naming pattern lemma_<P>_equiv
2. The Haskell type signature  
3. The function definition with `===` steps  
4. End with `* QED`
5. Please put your proof between ```haskell and  ```
No extra text, no additional comments.
Your answer must match the example format exactly, without trailing whitespace or newlines outside the code block.                                     

\end{lstlisting}

\subsection{Lemma SEQ Proof User Prompt}
\begin{lstlisting}
{error_msg_section}
                                      
{equiv_code}

------------------------------------------------------------

Your task: Produce the proof of equivalence for the following function:
`{func_name_p} x == {func_name_q} x` for all `x`.  

```haskell
{{-@ LIQUID "--reflection" @-}}
{{-@ LIQUID "--ple" @-}}
module Equiv where
import Language.Haskell.Liquid.ProofCombinators

{{-@ reflect {func_name_p} @-}}
{program_p_content}

{{-@ reflect {func_name_q} @-}}
{program_q_content}

-- Your complete proof of equivalence
/* PROOF BODY HERE */
```
<think>
\end{lstlisting}

\subsection{Alice SINQ System Prompt}
\begin{lstlisting}
You are an expert Haskell programmer. Your task is to generate a semantically inequivalent variant of a given Haskell program, which means that there must exist at least a diverging input example such that the original program and your program either produce different outputs or exceptions, or one halts and the other one does not halt.
You must also provide a diverging input, which is a valid input for both programs, but on which they produce different outputs.
                              
A good inequivalent program `Q` should be subtly different from `P`.
A good diverging input `x` should be simple and clearly demonstrate the semantic difference between `P` and `Q`.

The original program and your program will be used in a test to evaluate the skill of an expert Haskell programmer who will have to produce a diverging example (not necessarily the same as yours), so make sure that the difference you introduce are not very easy to understand. 
You will be given a difficulty level from 0 (easiest) to 10 (hardest) to target. E.g. difficulty level 0 means that an expert computer scientist in the bottom decile or above should be able to find a diverging example, difficulty level 9 means that only an expert computer scientist in the top decile should be able to find a diverging example, and difficulty level 10 means that only the top 0.01 or less of expert Haskell programmer should be able to find a diverging example.                                 

First, think step-by-step and write down your analysis of program `P` and your strategy for creating an inequivalent program `Q`. Enclose this reasoning within `<think>` and `</think>` tags.
After the thinking block, the final answer could only be in the following format, without any additional explanation or context.

Final output MUST be exactly: 
Generated Program `Q`:
```haskell
<Your generated Haskell code for `Q`>
```

Diverging Input `x`:
```
<The diverging input `x`>
```
\end{lstlisting}

\subsection{Alice SINQ User Prompt}\label{app:Alice_SINQ_User_Prompt}
\begin{lstlisting}
Difficulty level: {difficulty_level}
Original program `P`:
```haskell
{program}
```

<think>
\end{lstlisting}

\subsection{Bob System Prompt}
\begin{lstlisting}
You are an expert Haskell programmer. You are given two Haskell programs, `P` and `Q`.
Your task is to determine if they are semantically equivalent.
Use the following format to respond:
# Equivalent?
Yes or No

If the programs are equivalent, respond with your thought process and a final output with: 
# Equivalent?
Yes

If they are inequivalent, respond with your thought process and a final output with:
# Equivalent?
No
\end{lstlisting}
\subsection{Bob User Prompt}
\begin{lstlisting}
Program `P`:
```haskell
{program_p}
```

Program `Q`:
```haskell
{program_q}
```
<think>
\end{lstlisting}

\subsection{Alice SEQ Difficulty Prediction System Prompt}
\begin{lstlisting}
Difficulty level: Any
Program P
```haskell
{program_p}
```
The program Q below is semantically equivalent to the original program P, where the equivalence is due to the fact that the two programs have the same behavior on all inputs.
Program Q
```haskell
{program_q}
```
\end{lstlisting}
\subsection{Alice SINQ Difficulty Prediction System Prompt}
\begin{lstlisting}
Difficulty level: Any
program P
```haskell  
{program_p}
```
The program Q below is semantically inequivalent to the original program P, where the inequivalence is due to the fact that the two programs have different behavior on some inputs.
Program Q
```haskell
{program_q}
```
\end{lstlisting}
\subsection{Alice Difficulty Prediction User Prompt}
\begin{lstlisting}
Predict the difficulty level of the instance of Program Q compared to Program P. Just write \"Difficulty level: D\" where D is your prediction, do not write anything else.
\end{lstlisting}

\section{Main Experiment Setup}\label{app:main_exp_setup}

\subsection{Model Configs and Experiment Settings}

We employ \texttt{DeepSeek-R1-Distill-Qwen-7B} as the base model. The decoding parameters are configured as: temperature $T = 0.6$, top-$p = 0.95$, top-$k = 20$, presence penalty $= 1.5$ (Applied to reduce repetition in outputs from smaller reasoning models~\citep{mahaut2025repetitions}), and a context length of 32,768 tokens.

In $E_0$, Alice is configured with a 50/50 probability of playing either the SEQ or SINQ game (Section~\ref{sec:branching}), with the difficulty level set as 10 in her prompts (Appendix~\ref{app:Alice_SEQ_User_Prompt} and~\ref{app:Alice_SINQ_User_Prompt}). Bob serves solely as an evaluator and difficulty scorer, and he is not fine-tuned iteratively but only once after all rounds are complete. 

\subsection{Dataset Size and Difficulty Threshold}
Due to computational constraints, each experimental run uses a maximum of 500 Haskell reference programs\footnote{A main run with fine-tuning takes about 3 days on 4 NVIDIA L40S GPUs.}, i.e., $|\mathcal{D}| = P = 500$. The difficulty threshold $\tau$ is set to $3$ (rather than the default $5$) to increase the number of training examples available to Alice. For future experiments with greater resources, it is recommended to increase $|\mathcal{D}|$ and restore $\tau = 5$ to more closely simulate a fully adversarial setting.

\subsection{Fine-Tuning Protocol}
After each self-play round, Alice is fine-tuned on the combination of newly generated instances and the retained examples from all previous rounds. More importantly, each round’s fine-tuning is initialized from the original base model rather than from the fine-tuned adapter checkpoint obtained in the previous round. This strategy is adopted to avoid bias accumulation across iterations and to maintain stable and unbiased difficulty estimation. 

During generation in round $i+1$, Alice uses the LoRA learned in round $i$, but fine-tuning for that round still begins from the base model.

\subsection{Experimental Duration}
The main run comprises 7 self-play rounds. In each round, Alice is fine-tuned according to the above protocol. Bob undergoes a single fine-tuning step after all 7 rounds are complete. For both Alice and Bob, the fine-tuning minimises the loss only on the model’s own outputs.

\section{Rationale and Derivations for the SEQ-SINQ Comparison Experiment}\label{app:com_exp}

\subsection{Fairness Criterion and Derivation}

Let $P$ be the number of reference programs attempted per round, $p\in[0,1]$ the probability of playing SEQ (so $1-p$ for SINQ), and let $r_{\text{SEQ}}, r_{\text{SINQ}}$ denote the verification yields (probability that an attempt becomes a verified training example). By linearity of expectation, the expected number of verified examples per round is
$$
K(P,p) \;=\; P\bigl(p\,r_{\text{SEQ}} + (1-p)\,r_{\text{SINQ}}\bigr).
$$
To isolate the effect of SEQ vs. SINQ unbalanced volume size, we match  the expected count $K$ of verified pairs across arms. 

\subsection{Yield Estimation from the Main Run \texorpdfstring{$E_0$}{E0}}

From the 7-round of main experiment $E_0$, $|\mathcal{D}|=P=500$ (per-round attempts $\approx 250$ SEQ and $250$ SINQ):

\begin{table}[h!]
\centering
\small
\resizebox{\columnwidth}{!}{%
\begin{tabular}{l|c|c|c}
\toprule
\textbf{Game} & \textbf{Total Validated} & \textbf{Attempts} & \textbf{Yield ($\hat r$)} \\
\midrule
SEQ & 34 & 1,750 & 1.94\% \\
SINQ & 903 & 1,750 & 51.6\% \\
\bottomrule
\end{tabular}%
}
\caption{Validation yields from $E_0$. Validated counts by round were: SEQ ($1,2,4,6,3,9,9$) and SINQ ($130,139,124,128,131,138,113$).}
\label{tab:yield_estimation}
\end{table}

\subsection{Predicting Supervision for Each Arm}

To achieve a balanced distribution of verified training examples (approx. 50/50 split), we configured regime $E_2$ with a mix of 96\% SEQ and 4\% SINQ. With a reference budget of $P=500$, the estimated verified yields per round are:
\begin{align}
    N_{\text{SEQ}} &\approx 500 \cdot 0.96 \cdot 0.0194 \approx 9.3 \\
    N_{\text{SINQ}} &\approx 500 \cdot 0.04 \cdot 0.516 \phantom{0} \approx 10.3
\end{align}
Totaling these yields using $K(P,p)$ results in an expected $K \approx 19.6$ verified examples per round, achieving the desired parity between the two supervision signals.
\subsection{Matched comparator for \texorpdfstring{$E_2$}{E2}}

Choose $P_{\text{$E_3$}}$ in a 100\% SINQ run so that
\begin{equation}
\begin{split}
    P_{E_3} \cdot r_{\text{SINQ}} &= K_{E_2} \\
    \Rightarrow P_{E_3} &= \frac{19.632}{0.516} = 38.04 \approx 40
\end{split}
\label{eq:p_e3_derivation}
\end{equation}

\subsection{What this controls, and what it does not}

Matching $K$ ensures each arm receives similar amounts of verified training, so outcome differences are attributable to the type of game (presence/absence of SEQ) rather than volume. However, residual differences may still arise from the difficulty distribution of accepted items and class-imbalance within Bob’s updates.

\subsection{Comparison Objectives}

The experimental design supports the following four key pairwise comparisons:

\begin{itemize}
    \item $E_0 \text{ vs. } E_1$: Under fixed compute resources (constant $P$), does a mixed 50\% SEQ / 50\% SINQ regime yield greater performance improvements than a pure SINQ regime similar to Miceli et al. setting~\citep{miceli2025program}, given that the latter produces substantially more verified training examples due to higher verification success rates in SINQ?
    \item $E_0 \text{ vs. } E_2$: With fixed compute resources, does shifting towards a more balanced number of verified SEQ and SINQ examples, at the cost of reducing the overall number of verified training pairs, lead to improved evaluator performance compared to a SINQ-dominated main experiment?
    \item $E_1 \text{ vs. } E_2$: Under the same compute constraints, does a pure SINQ regime (maximising verified training pairs) outperform a more balanced SEQ–SINQ dataset that sacrifices data volume for greater semantic diversity?
    \item $E_2 \text{ vs. } E_3$ When the number of verified training pairs is held constant, does the inclusion of SEQ supervision yield measurable performance benefits over an SINQ-only regime?
\end{itemize}

\subsection{Summary of experiments}

\begin{itemize}
    \item $E_0$ (main): 50/50, $P=500$.
    \item $E_1$: 100\% SINQ, $P=500$.
    \item $E_2$: 96\% SEQ / 4\% SINQ, $P=500$.
    \item $E_3$ (matched to $E_2$): 100\% SINQ, $P=40$  to equalize expected verified pairs with $E_2$.
\end{itemize}

This suite isolates the causal effect of adding SEQ supervision under controlled total verified pairs, enabling a fair assessment of whether SEQ contributes beyond SINQ alone.

\onecolumn

\section{Experiment Results}\label{app:exp_result}

\begin{table}[h!]
\centering
\begin{tabularx}{\linewidth}{
    >{\raggedright\arraybackslash}p{4cm}  % First column wider
    >{\raggedright\arraybackslash}p{2cm}  % Second column (Metrics)
    >{\centering\arraybackslash}p{2cm}    % E0
    >{\centering\arraybackslash}p{2cm}    % E1
    >{\centering\arraybackslash}p{2cm}    % E2
    >{\centering\arraybackslash}p{2cm}    % E3
}
\toprule
& & \multicolumn{4}{c}{\textbf{Score (\%)}} \\
\cmidrule(lr){3-6}
\textbf{Benchmark} & \textbf{Metric}
& $\mathbf{E_0}$ & $\mathbf{E_1}$ & $\mathbf{E_2}$ & $\mathbf{E_3}$ \\
\midrule
HumanEval (Bob) & Pass@1 & \textbf{26.4} & 25.8 & 22.7 & 22.0 \\
HumanEval (Alice) & Pass@1 & 26.3 & \textbf{28.1} & 21.8 & 18.4 \\
MBPP (Bob) & Pass@1 & 36.9 & \textbf{38.8} & 34.7 & 32.4 \\
MBPP (Alice) & Pass@1 & 34.3 & \textbf{39.3} & 32.9 & 28.0 \\
EquiBench (OJ\_A) & Acc & \textbf{57.4} & 54.0 & 55.5 & 53.3 \\
EquiBench (OJ\_A) & F1 & \textbf{50.7} & 44.4 & 48.9 & 45.1 \\
EquiBench (OJ\_V) & Acc & \textbf{69.8} & 61.4 & 63.2 & 65.8 \\
EquiBench (OJ\_V) & F1 & \textbf{69.7} & 58.0 & 61.5 & 64.7 \\
EquiBench (OJ\_VA) & Acc & \textbf{55.9} & 54.5 & 54.7 & 53.7 \\
EquiBench (OJ\_VA) & F1 & \textbf{47.5} & 45.4 & 47.1 & 45.2 \\
EquiBench (STOKE) & Acc & \textbf{51.8} & 50.9 & 51.0 & 50.8 \\
EquiBench (STOKE) & F1 & \textbf{18.3} & 15.4 & 16.8 & 15.9 \\
EquiBench (TVM) & Acc & 50.1 & 50.3 & 48.8 & \textbf{50.8} \\
EquiBench (TVM) & F1 & 38.3 & \textbf{39.9} & 33.5 & 33.9 \\
EquiBench (DCE) & Acc & 51.9 & \textbf{52.6} & 51.5 & 51.3 \\
EquiBench (DCE) & F1 & \textbf{47.7} & 45.4 & 45.2 & 44.4 \\
PySecDB & Acc & \textbf{68.8} & 67.3 & 68.7 & 68.7 \\
PySecDB & F1 & \textbf{54.0} & 51.9 & 53.9 & 53.6 \\
CodeXGlue & Acc & \textbf{50.5} & 49.2 & 49.9 & 48.9 \\
CodeXGlue & F1 & 43.2 & 44.0 & 44.1 & \textbf{48.5} \\

\bottomrule
\end{tabularx}
\caption{Benchmark Performances over all experiments. Averages over 16 samples.}
\label{tab:benchmark_perf_all_exp}

\begin{tabularx}{\linewidth}{
    >{\centering\arraybackslash}p{4cm}  % Second column (Metrics)
    >{\centering\arraybackslash}p{2cm}    % E0
    >{\centering\arraybackslash}p{2cm}    % E1
    >{\centering\arraybackslash}p{2cm}    % E2
    >{\centering\arraybackslash}p{2cm}    % E3
}
\toprule
& \multicolumn{4}{c}{\textbf{Count}} \\
\cmidrule(lr){2-5}
 \textbf{Game}
& $\mathbf{E_0}$ & $\mathbf{E_1}$ & $\mathbf{E_2}$ & $\mathbf{E_3}$ \\
\midrule
SEQ & 34 & -- & 62 & -- \\
SINQ & 903 & 1,839 & 78 & 156 \\
\midrule
Total & 937 & 1,839 & 140 & 156 \\
\bottomrule
\end{tabularx}
\caption{Total verified generation from Alice of all experiments over 7 rounds.}
\label{tab:basic_metric_perf_all_exp}
\end{table}

\twocolumn
\section{Extended Regime Results}\label{app:eval_result}

\subsection{Experiment \texorpdfstring{$E_0$}{E0}}\label{app:eval_result_e0}

Most of the evaluation results are included in the main report. 

\subsubsection{Difficulty trajectories for both task types}\label{app:difficulty_trajectory}

\begin{figure}[htbp] % or [htbp]
    \centering
    % First subfigure
    \begin{subfigure}{0.5\textwidth}
        \centering
        \includegraphics[width=\linewidth]{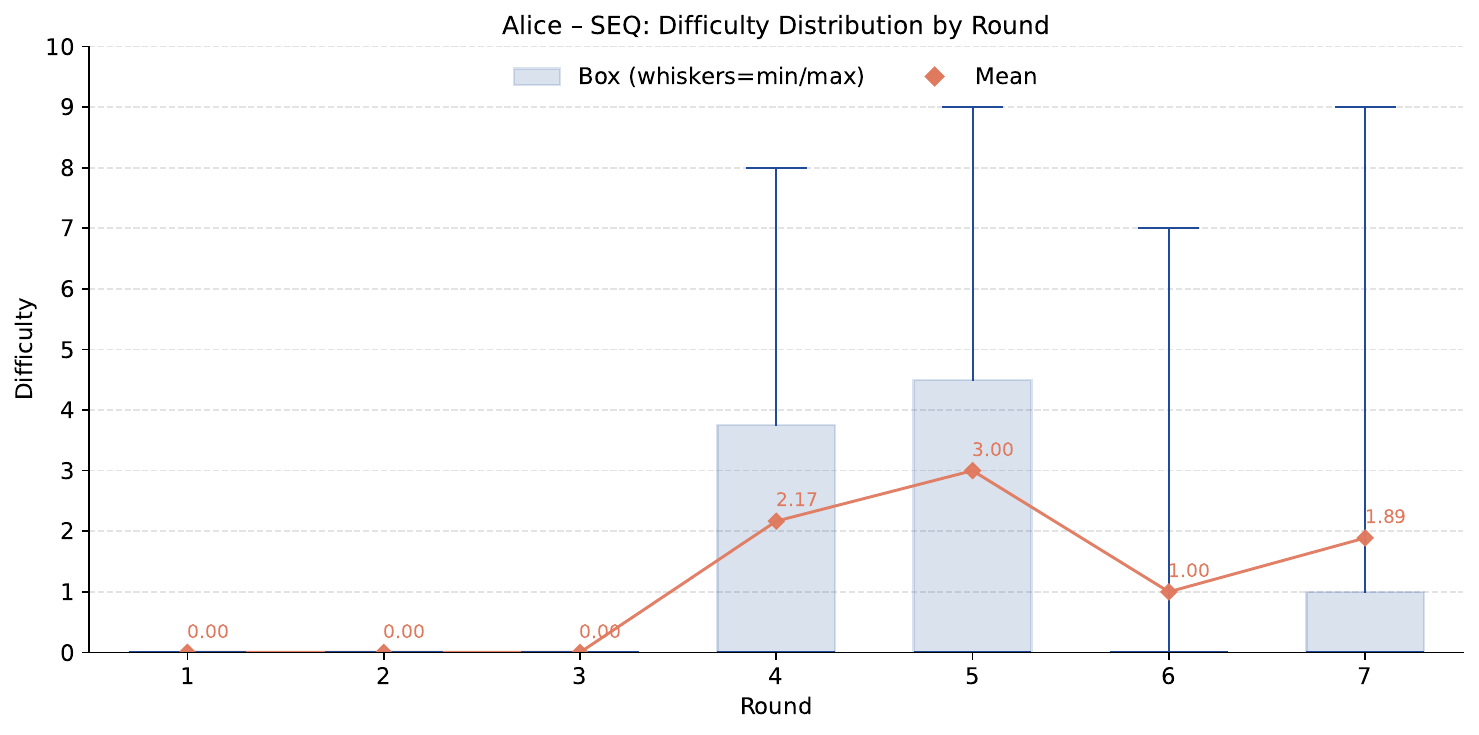}
    \end{subfigure}
    % Second subfigure
    \begin{subfigure}{0.5\textwidth}
        \centering
        \includegraphics[width=\linewidth]{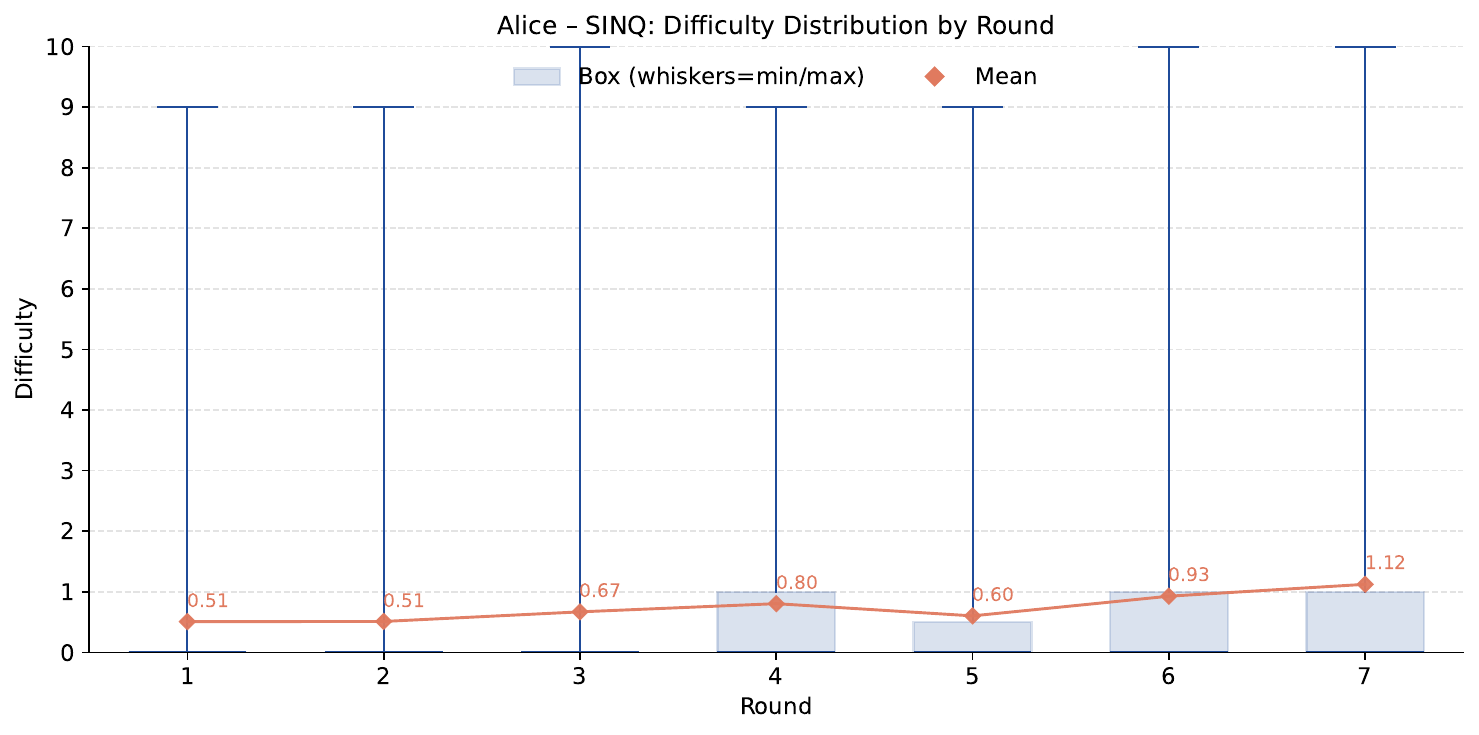}
    \end{subfigure}

    \caption{Mean and box-whiskey plots of the difficulty scores of Alice's generated instances by round. Top: SEQ Game. Bottom SINQ Game. }
    \label{fig:box_diff_by_round}
\end{figure}

When we inspect the difficulty trajectories for both task types in Figure~\ref{fig:box_diff_by_round}, they move upward across rounds.

\begin{itemize}
    \item For SEQ, although the sample size is not statistically significant, its mean difficulty score jumps from near $0.0$ in early rounds to $2.0 - 3.0$ in later ones, signaling that Alice manages to construct are non-trivial from Bob’s perspective.
    \item For SINQ, its averages being greater than the medians ($0.0$ for all rounds) indicates a right‐skewed distribution. This suggests that while most generated instances are of lower difficulty, Alice occasionally produces much harder examples that significantly influence the mean. Starting in round 4, the magnitude of the difference between the mean and median becomes larger, and by round 7 the mean even exceeds the upper quartile. Therefore, Alice is producing increasingly harder examples over successive rounds.
\end{itemize}

Both SEQ and SINQ setups show rising difficulty scores across rounds, indicating that Alice improves in generating more challenging instances for Bob.

\subsubsection{Alice's HumanEval (Haskell) Performance}\label{app:e0_alice_Humaneval}

\begin{figure}[H]
    \centering
    \includegraphics[width=\linewidth]{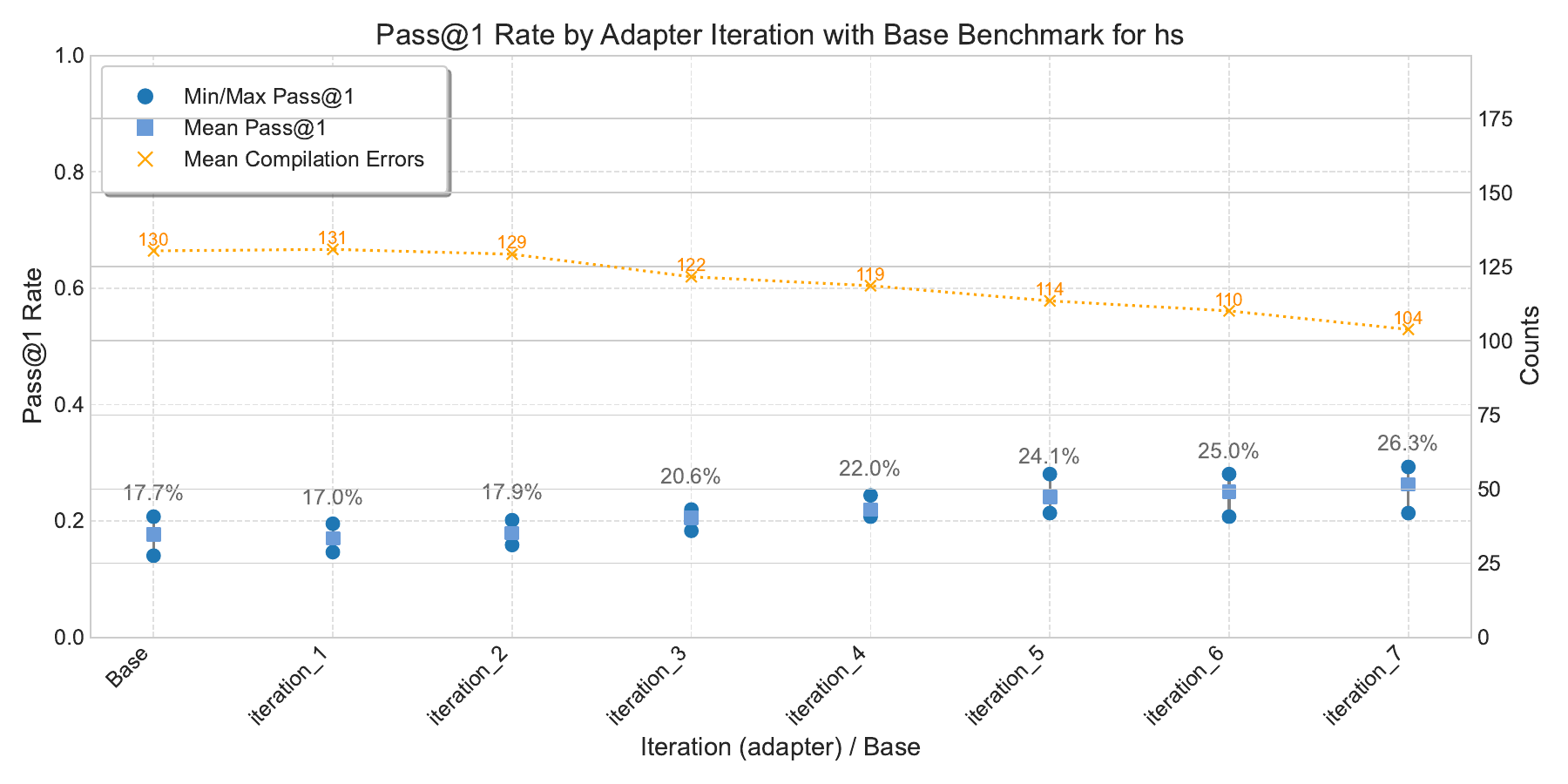}
    \caption{Alice's HumanEval results. Averages over 16 trials.}
    \label{fig:alice_humaneval_base_perf}
\end{figure}

\subsubsection{Alice's MBPP (Haskell) Performance}\label{app:e0_alice_MBPP}

\begin{figure}[H]
    \centering
    \includegraphics[width=\linewidth]{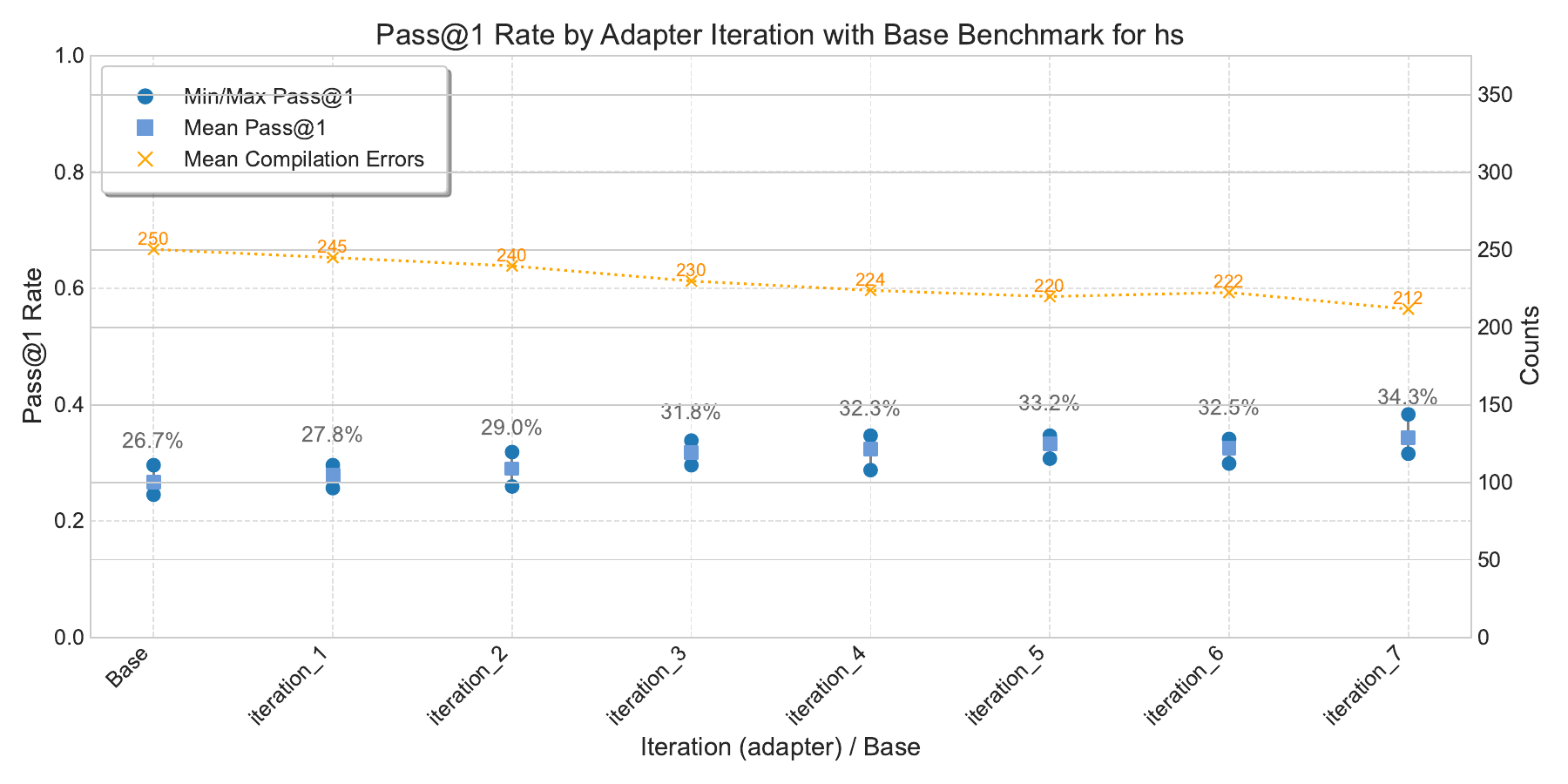}
    \caption{Alice's MBPP results. Averages over 16 trials.}
    \label{fig:alice_humaneval_base_perf_MBPP}
\end{figure}

\subsection{Experiment \texorpdfstring{$E_1$}{E1}}

\subsubsection{Alice's HumanEval (Haskell) Performance}

\begin{figure}[H]
    \centering
    \includegraphics[width=\linewidth]{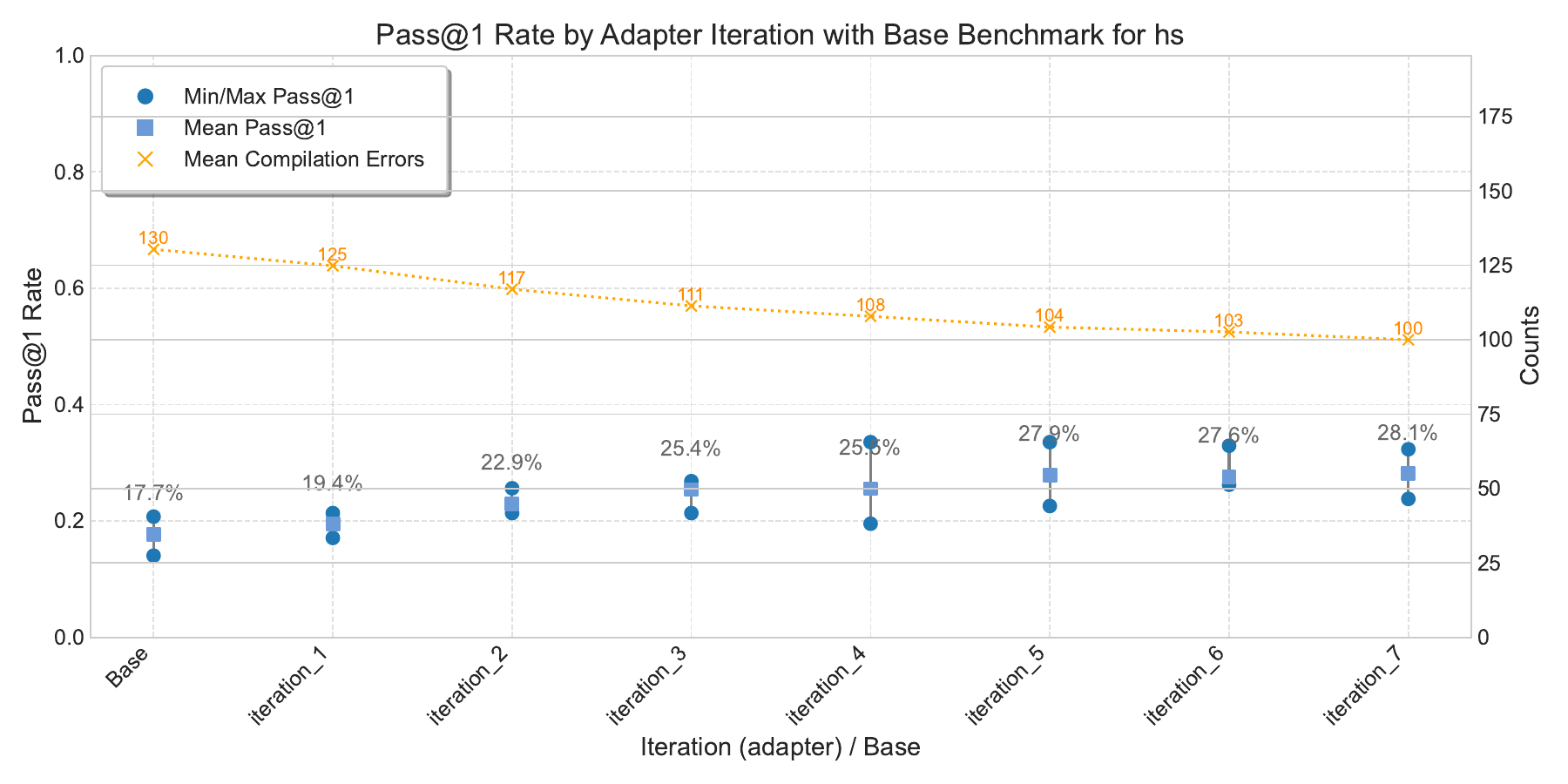}
    \caption{Alice's HumanEval results. Averages over 16 trials.}
    \label{fig:e1_alice_humaneval_base_perf}
\end{figure}

\subsubsection{Alice's MBPP (Haskell) Performance}

\begin{figure}[H]
    \centering
    \includegraphics[width=\linewidth]{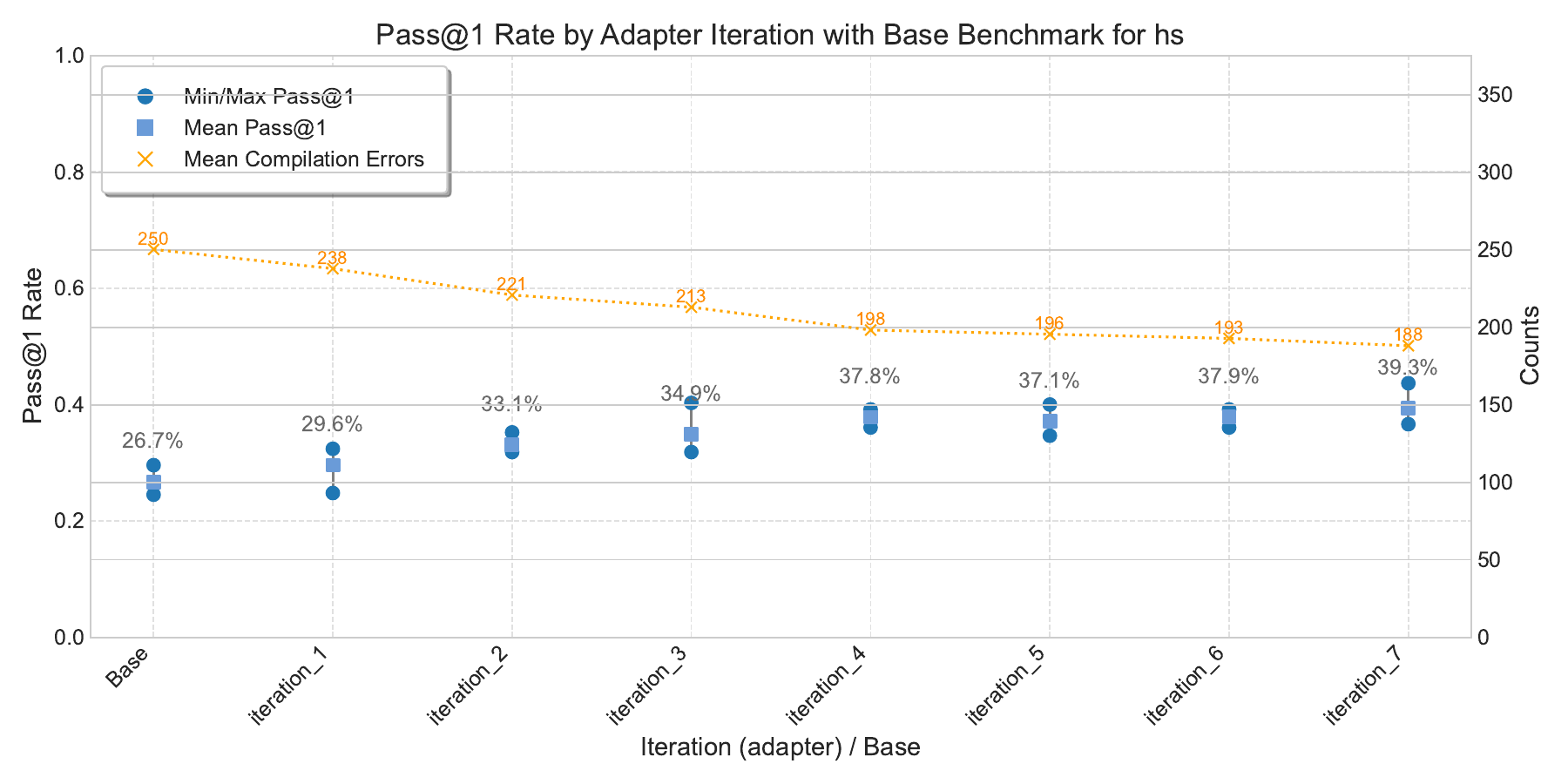}
    \caption{Alice's MBPP results. Averages over 16 trials.}
    \label{fig:e1_alice_humaneval_base_perf_MBPP}
\end{figure}

\subsubsection{Alice's SEQ vs SINQ Validated Generation Counts}

\begin{figure}[H]
    \centering
    \includegraphics[width=\linewidth]{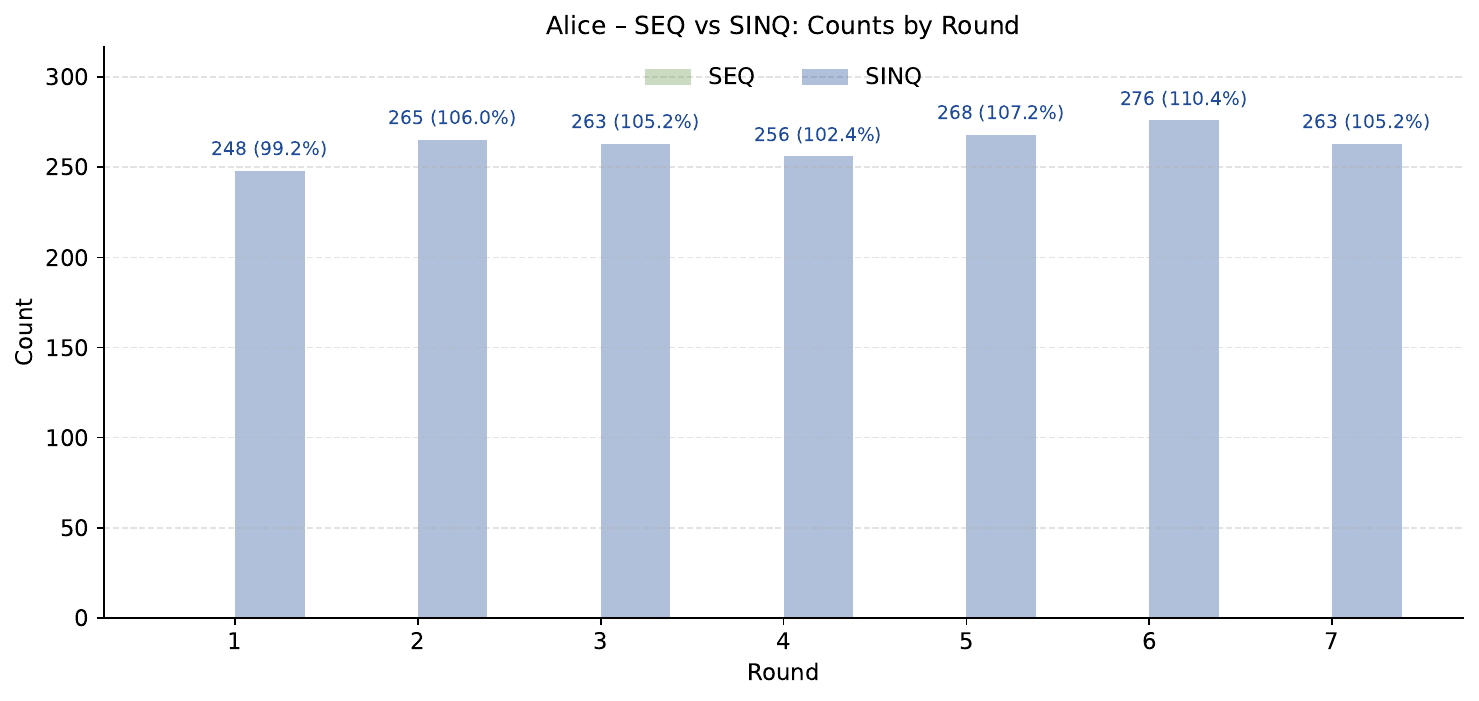}
    \caption{SEQ vs SINQ Validated Generation Counts.}
    \label{fig:e1_seq_sinq_counts_side_by_side_perf}
\end{figure}

\subsection{Experiment \texorpdfstring{$E_2$}{E2}}

\subsubsection{Alice's HumanEval (Haskell) Performance}

\begin{figure}[H]
    \centering
    \includegraphics[width=\linewidth]{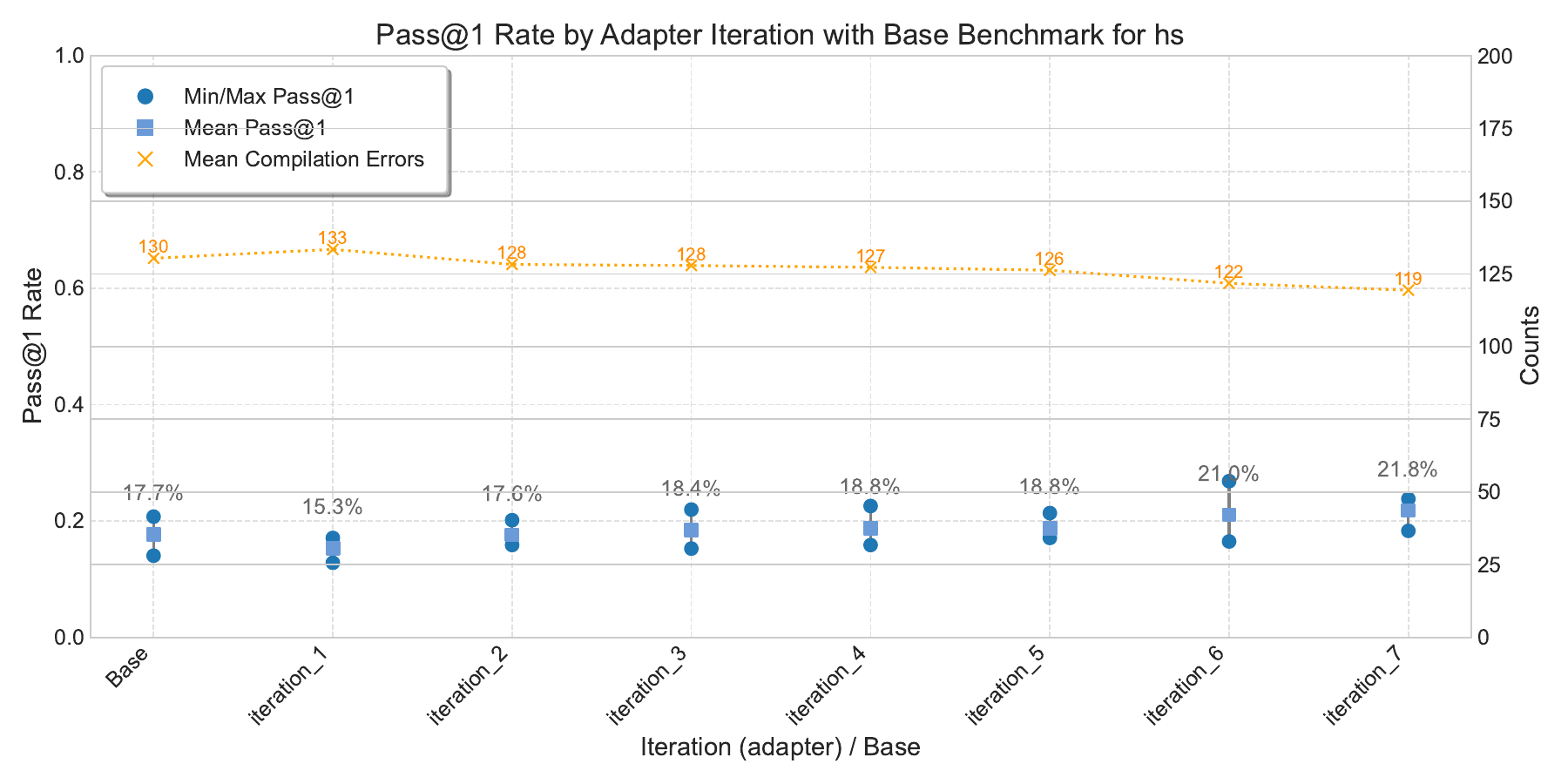}
    \caption{Alice's HumanEval results. Averages over 16 trials.}
    \label{fig:e2_alice_humaneval_base_perf}
\end{figure}

\subsubsection{Alice's MBPP (Haskell) Performance}

\begin{figure}[H]
    \centering
    \includegraphics[width=\linewidth]{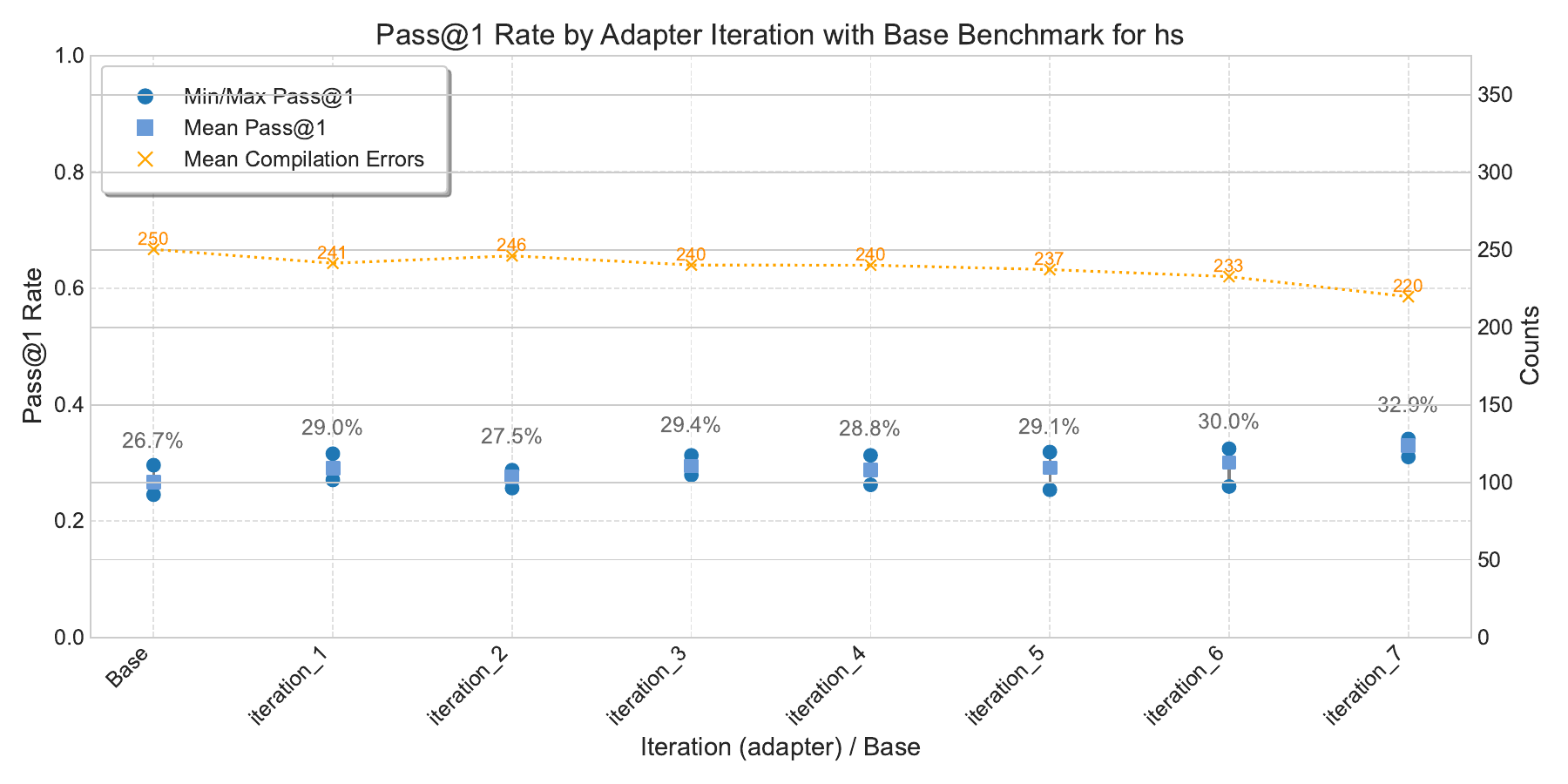}
    \caption{Alice's MBPP results. Averages over 16 trials.}
    \label{fig:e2_alice_humaneval_base_perf_MBPP}
\end{figure}

\subsubsection{Alice's SEQ vs SINQ Validated Generation Counts}

\begin{figure}[H]
    \centering
    \includegraphics[width=\linewidth]{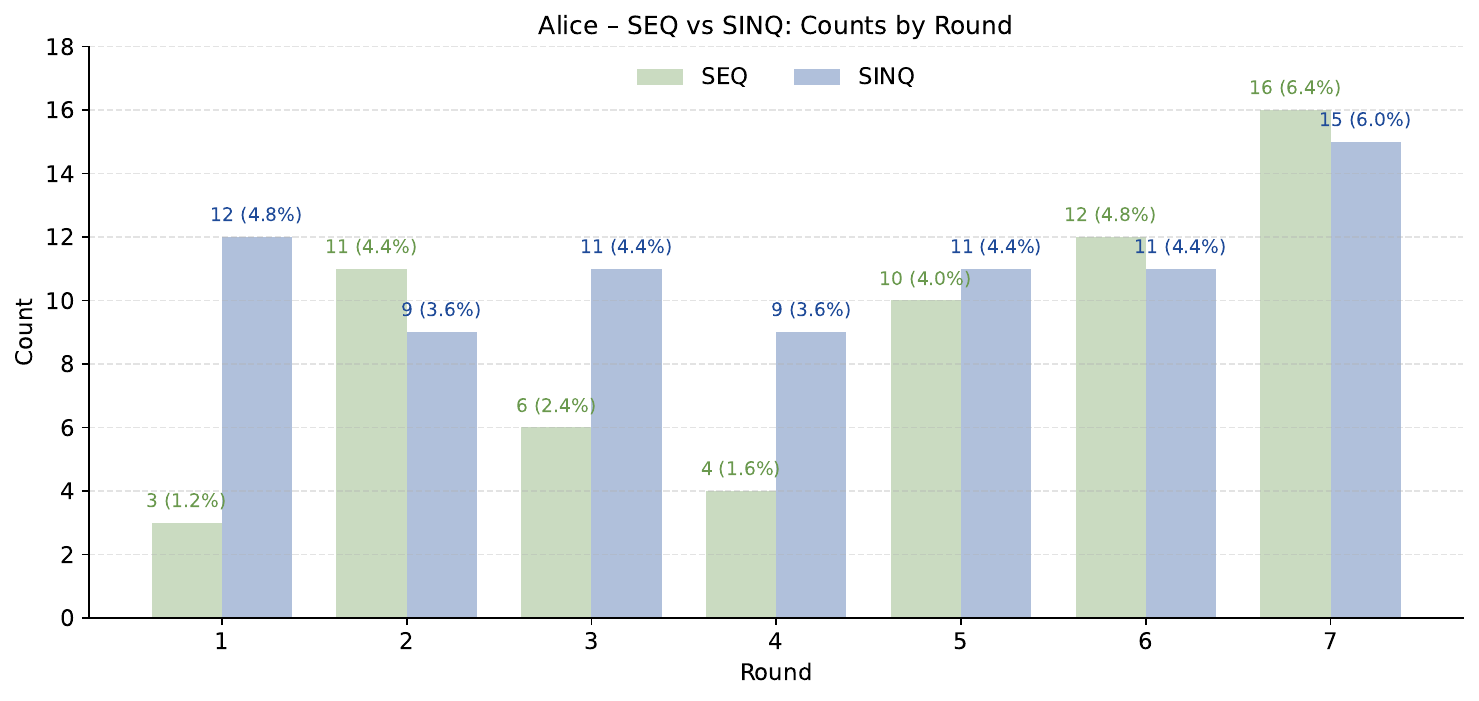}
    \caption{SEQ vs SINQ Validated Generation Counts.}
    \label{fig:e2_seq_sinq_counts_side_by_side_perf}
\end{figure}

\subsection{Experiment \texorpdfstring{$E_3$}{E3}}

\subsubsection{Alice's HumanEval (Haskell) Performance}

\begin{figure}[H]
    \centering
    \includegraphics[width=\linewidth]{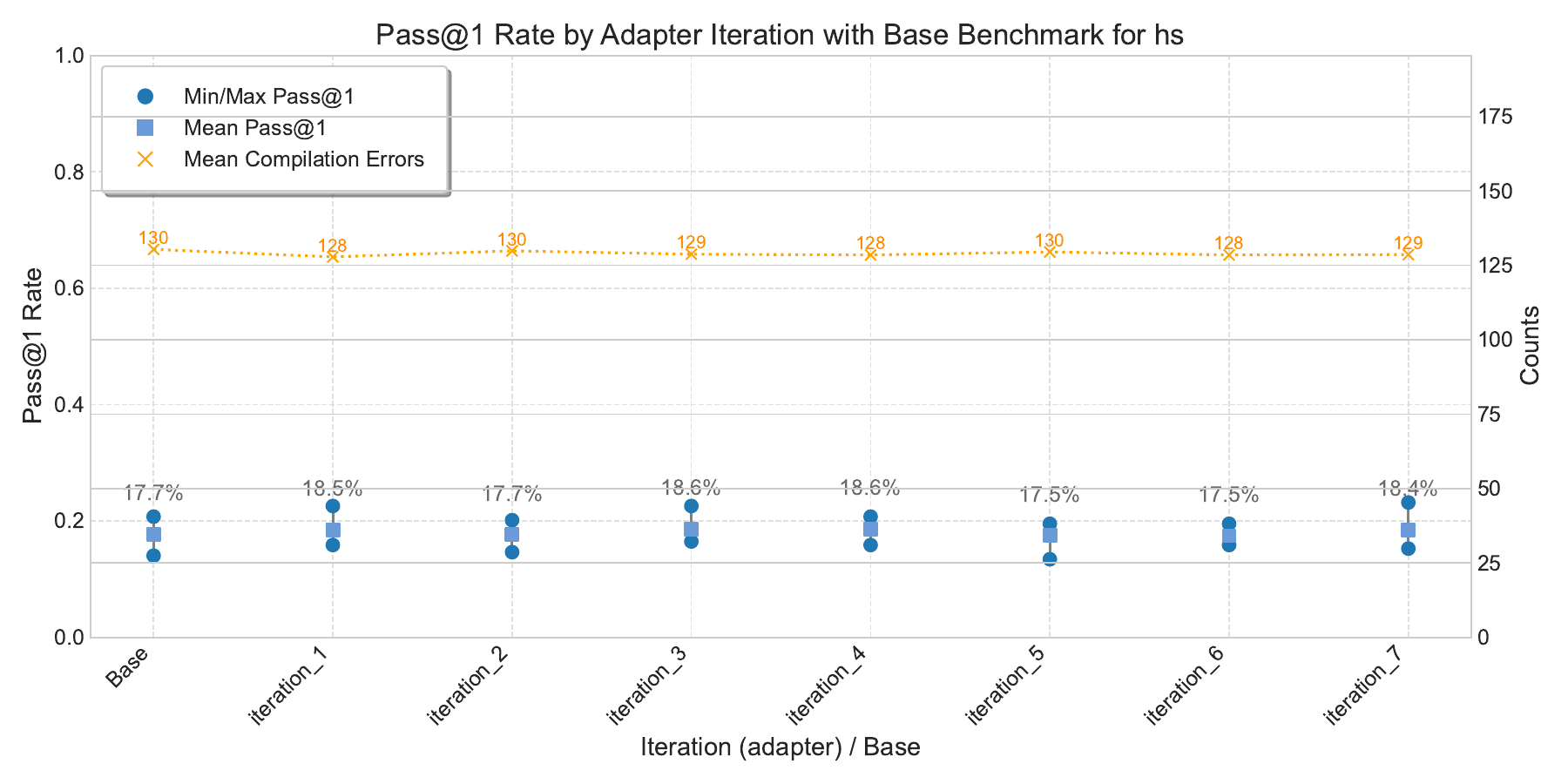}
    \caption{Alice's HumanEval results. Averages over 16 trials.}
    \label{fig:e3_alice_humaneval_base_perf}
\end{figure}

\subsubsection{Alice's MBPP (Haskell) Performance}

\begin{figure}[H]
    \centering
    \includegraphics[width=\linewidth]{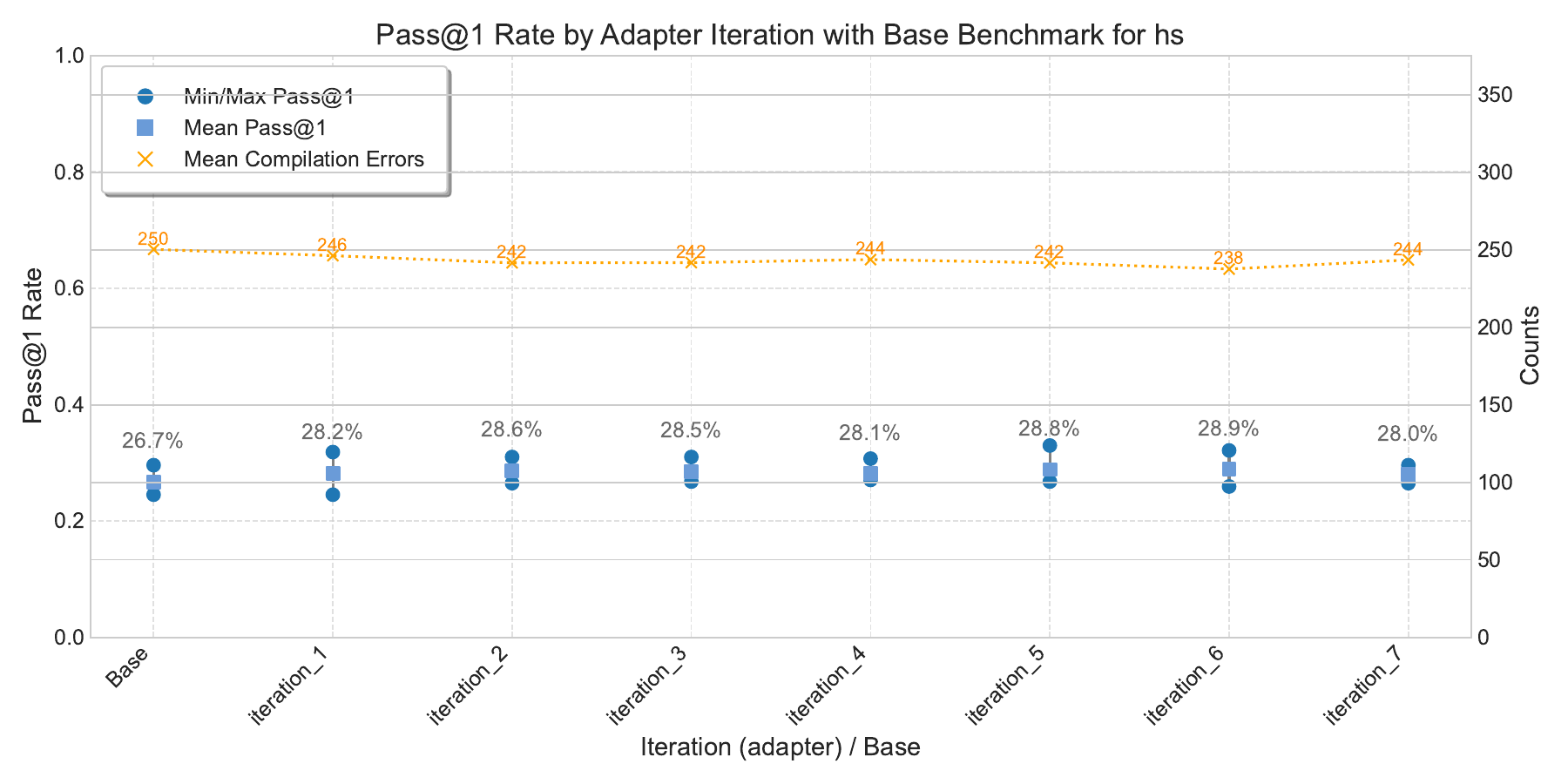}
    \caption{Alice's MBPP results. Averages over 16 trials.}
    \label{fig:e3_alice_humaneval_base_perf_MBPP}
\end{figure}

\subsubsection{Alice's SEQ vs SINQ Validated Generation Counts}

\begin{figure}[H]
    \centering
    \includegraphics[width=\linewidth]{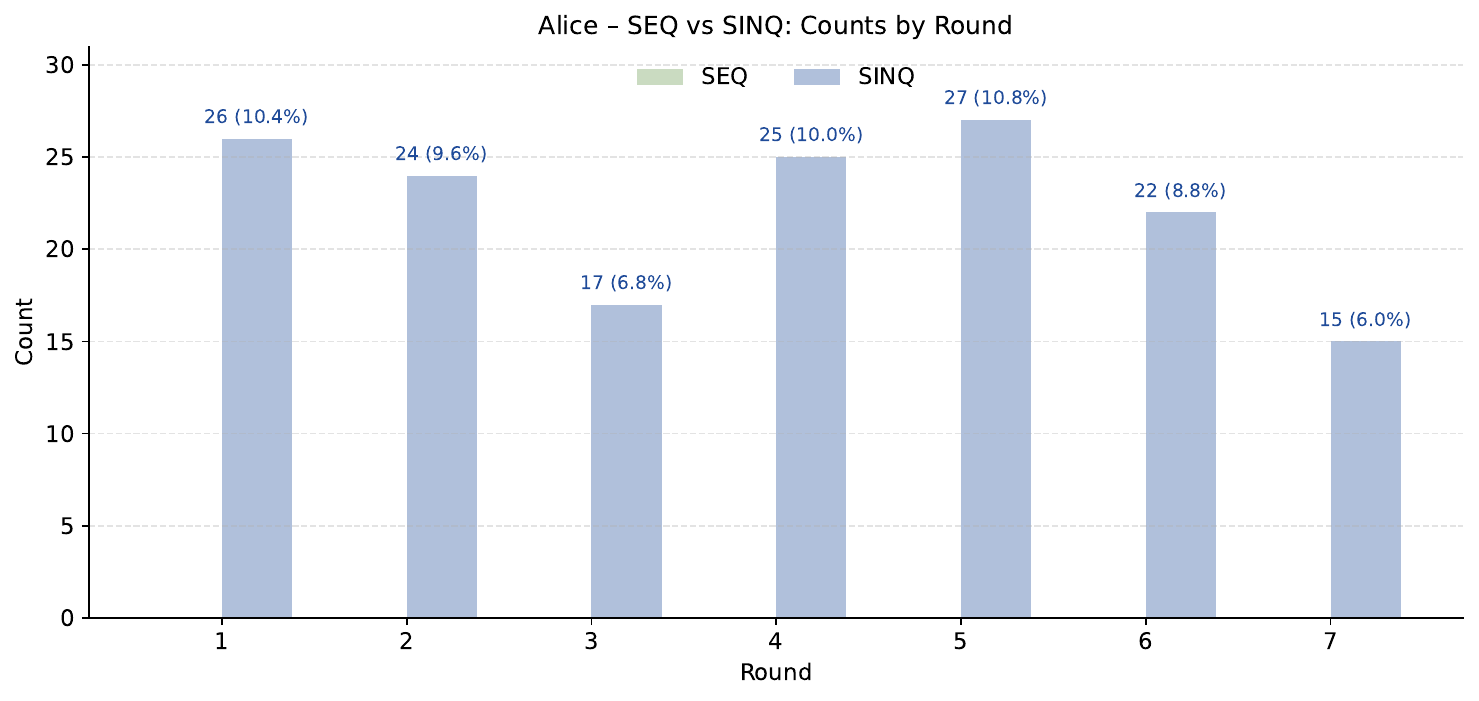}
    \caption{SEQ vs SINQ Validated Generation Counts.}
    \label{fig:e3_seq_sinq_counts_side_by_side_perf}
\end{figure}

\subsubsection{\texorpdfstring{$E_2$}{E2} vs \texorpdfstring{$E_3$}{E3} Validated Generation Counts}

\begin{figure}[H]
    \centering
    \includegraphics[width=\linewidth]{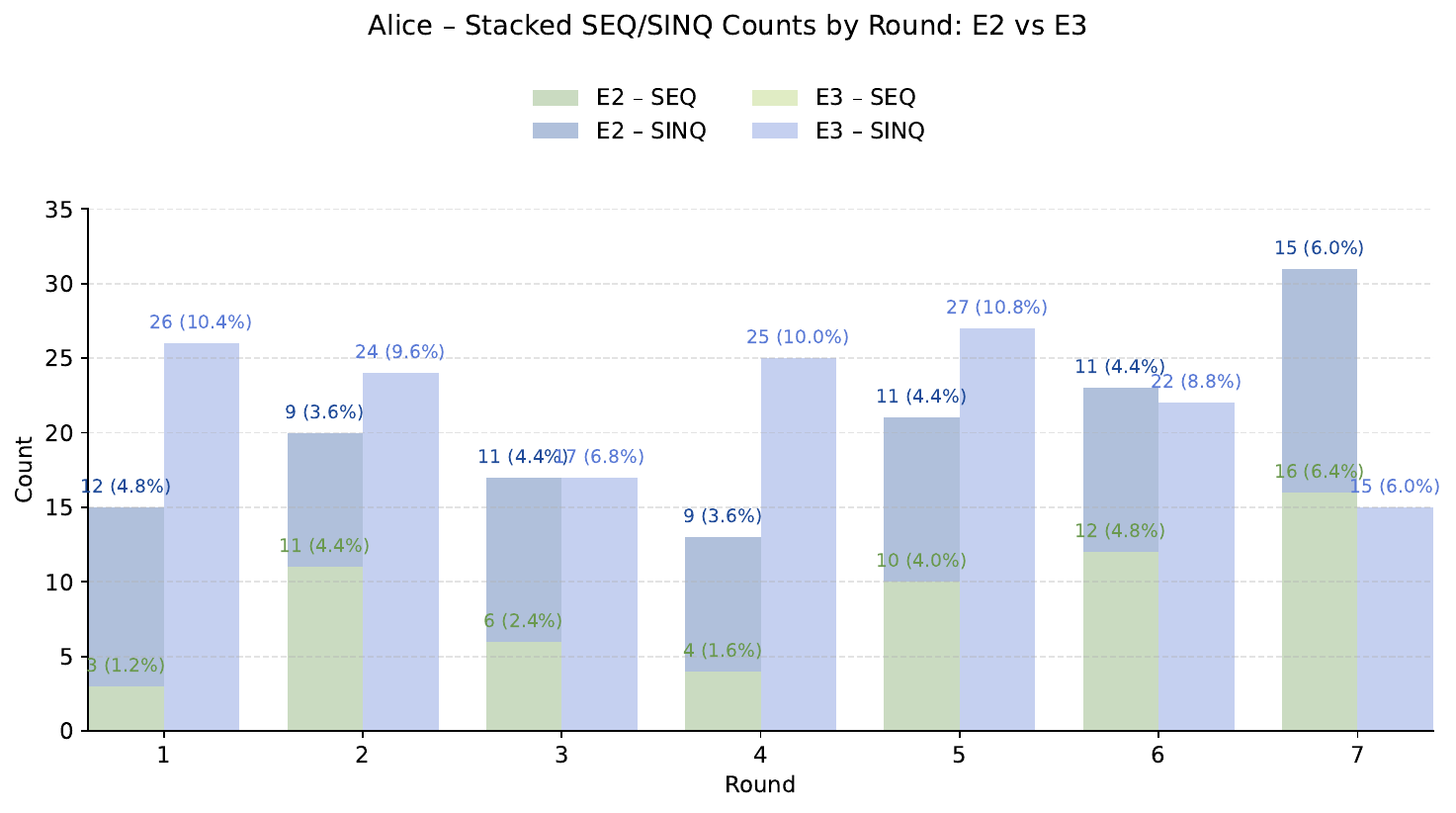}
    \caption{Validated Generation Counts of both experiment, indicating the effort of controlling $P$ shown in Appendix~\ref{app:com_exp} in order to achieve an equal number of verified generation.}
    \label{fig:e3_stacked_seq_sinq_counts_comparison_perf}
\end{figure}

\end{document}